\documentclass{article}

 \usepackage[preprint]{neurips_2026}

\usepackage[utf8]{inputenc} 
\usepackage[T1]{fontenc}    
\usepackage{hyperref}       
\usepackage{url}            
\usepackage{booktabs}       
\usepackage{amsfonts}       
\usepackage{nicefrac}       
\usepackage{microtype}      
\usepackage{xcolor}         

\usepackage{amsmath}
\usepackage{graphicx}
\usepackage{subcaption} 
\usepackage{float}
\usepackage{booktabs}   
\usepackage{makecell}   
\usepackage{comment}

\usepackage[table]{xcolor}
\usepackage{multirow}
\usepackage[dvipsnames, table]{xcolor}
\usepackage{booktabs}
\usepackage{float}

\title{Beyond Perplexity: A Geometric and Spectral Study of Low-Rank Pre-Training}

\author{%
\begin{tabular}{cc}
  \\
  \textbf{Namrata Shivagunde}\thanks{Corresponding author. Code: \url{https://github.com/NamrataRShivagunde/low-rank-geometry}} & \textbf{Vijeta Deshpande} \\
  \texttt{namratashivagunde@gmail.com} & \texttt{vijeta\_deshpande@student.uml.edu} \\[6pt]
  \textbf{Sherin Muckatira} & \textbf{Anna Rumshisky} \\
  \texttt{smuckati@cs.uml.edu} & \texttt{arum@cs.uml.edu} \\[6pt]
  \multicolumn{2}{c}{University of Massachusetts Lowell}
\end{tabular}
}

\begin{document}

\maketitle

\begin{abstract}
Pre-training large language models is dominated by the memory cost of storing full-rank weights, gradients, and optimizer states. Low-rank pre-training has emerged to address this, and the space of methods has grown rapidly. A central question remains open: do low-rank methods produce models that generalize comparably to full-rank training, or does the rank constraint fundamentally alter the solutions reached? Existing comparisons rely almost entirely on validation perplexity from single-seed runs, often carried forward from prior literature. Yet perplexity is a poor proxy for solution quality — two methods can match on perplexity while converging to very different loss landscape regions and internal representations. We close this gap by characterizing the solutions found by five low-rank pre-training methods — GaLore and Fira (memory-efficient optimizers), CoLA and SLTrain (architecture reparameterizations), and ReLoRA (adapter-style updates with periodic resets) — against full-rank training at three model scales (60M, 130M, 350M). We evaluate each along 16 metrics across four dimensions: 1-D loss landscape along random/top-K PCA directions, 1-D interpolation between checkpoints, spectral structure of the weights and learned updates, and activation similarity to full-rank training. We show that low-rank methods are not equivalent to full-rank training, nor to one another, even when validation perplexity is close. Full-rank training settles into sharper basin than low-rank methods along random directions, while the reverse holds for the top-1 PCA direction. Each method converges to a geometrically distinct basin. Low-rank activations increasingly diverge from full-rank in later layers as training progresses, with GaLore tracking full-rank most closely. Further, validation perplexity does not translate to downstream performance at every scale. Adding geometric and spectral metrics to it improves the prediction.


\end{abstract}
\section{Introduction}
\label{sec:intro}

Pre-training large language models (LLMs) is dominated by the memory cost of storing full-rank weights, gradients, and optimizer states. Low-rank pre-training has emerged as a way to reduce this cost while still approximating full-rank training. This field has evolved along three main directions. Optimizer-based methods such as GaLore~\citep{Zhao2024GaLoreML} leave the model intact and constrain only the gradient update to a low-rank subspace. Architecture-based methods such as CoLA~\citep{liu2025cola} replace dense linear layers with low-rank decompositions. Adapter-based methods such as ReLoRA~\citep{Lialin2023ReLoRAHT},
add low-rank adapters to  linear layers that are periodically merged into the weights and re-initialized.

Despite this growing body of work, evaluation practices in low-rank pretraining remain limited. Methods are benchmarked against one another using validation perplexity from a single training run with a single random seed \cite{Zhao2024GaLoreML, Lialin2023ReLoRAHT, liu2025cola, Han2024SLTrainAS, Loeschcke2024LoQTLR, zhou2024switchlora, Muhamed2024GrassCE, Shivagunde2024, li2025lost, miles2024velora}. While this is understandable given the significant compute costs of large-scale pretraining, it raises an important question: is validation perplexity alone a sufficient criterion for evaluating a low-rank pretraining method's quality? Earlier work \citep{Liu2022SamePL} shows that two models with the same pre-training loss can differ substantially on downstream tasks, and that loss-landscape flatness predicts this gap. Beyond validation perplexity, most prior work evaluates low-rank pre-training methods by using them as fine-tuning recipes. For example, GaLore~\citep{Zhao2024GaLoreML}, SLTrain~\citep{Han2024SLTrainAS}, LOST~\citep{li2025lost}, and VeLoRA~\citep{miles2024velora} are evaluated by fine-tuning RoBERTa on GLUE, while CoLA~\citep{liu2025cola} is evaluated by fine-tuning LLaMA-7B on downstream tasks. This measures how well a low-rank method fine-tunes a pre-trained model, but does not tell us how a model pre-trained with the low-rank method generalizes downstream. A few studies do fine-tune pre-trained models directly. For instance, Lialin et al.~\cite{Lialin2023ReLoRAHT} compare ReLoRA against full-rank on GLUE, Liu et al.~\cite{liu2025cola} compare a pre-trained CoLA against BERT, and Zhou et al.~\cite{zhou2024switchlora} compare full-rank, GaLore, and SwitchLoRA. These studies are narrow in scope, covering one or two methods at a single scale, and none span all three method families (optimizers, model architecture, adapters).

Only a handful of recent studies look beyond validation perplexity, and these are narrow in scope. FlatLoRA~\cite{flatlora} compares LoRA against full-rank training via loss landscape analysis, but only in the fine-tuning setting. Liu et al.~\cite{plnumber} study GaLore, LoRA, and full-rank training through the Polyak-Łojasiewicz (PL) condition, but their analysis is restricted to multi-layer perceptrons and does not transfer cleanly to transformers. Both studies cover only a subset of low-rank methods and rely on a single analytical lens.

%

We address this research gap with four contributions: 
\textbf{(1) A diagnostic framework for low-rank pre-training}, including a novel cross-method barrier metric (IMBH) that quantifies geometric separation between training algorithms — not just between seeds or epochs as in prior linear-mode-connectivity work.
\textbf{(2) Cross-family empirical findings} that no prior work establishes: low-rank methods occupy geometrically separated basins (IMBH grows monotonically), develop distinct spectral signatures (GaLore matches full-rank, CoLA don't), and produce divergent internal representations (CoLA orthogonal, Fira recovers only at last layer).
{\textbf{(3) Geometry-augmented predictor of downstream performance.}} The perplexity ranking of low-rank methods does not translate to zero-shot downstream performance, but augmenting perplexity with geometric and spectral features from our framework improves the prediction.
\textbf{(4) An open-source unified codebase.} We release a single pipeline integrating pre-training, metric computation, and downstream evaluation, designed for easy extension to new methods and metrics, enabling the community to study low-rank pre-training in more depth and design better methods.

\section{Related Work}
\label{sec:related-work}
\vspace{-0.3em}
\paragraph{Low-rank pre-training methods.}
Memory-efficient pre-training of LLMs has motivated a fast-growing family of low-rank methods, which broadly fall into three design categories. The first category wraps the optimizer, constraining the gradient update to a low-rank subspace while leaving the parameters at full rank. GaLore \citep{Zhao2024GaLoreML} projects gradients into a low-rank subspace, applies an Adam update there, and projects back to the full-rank parameter space. Fira \citep{Chen2024FiraCW} introduces a norm-based scaling that uses GaLore's low-rank Adam states to scale the full-rank gradient, recovering full-rank update directions at the same memory cost. A growing line of work extends this design idea along several axes, including Q-GaLore \citep{zhang2024q}, GaLore 2 \citep{Su2025GaLore2L}, GRASS \citep{Muhamed2024GrassCE}, GaLore-mini \citep{huang2024galore}, COAP \citep{xiao2025coap}, and FLORA \citep{hao2024flora}. The second category modifies the model itself, replacing dense weight matrices with low-rank decompositions. CoLA \citep{liu2025cola} restructures attention and MLP linear layers as bottleneck low-rank matrices with a non-linearity in between. SLTrain \citep{Han2024SLTrainAS} parameterizes weights as a sparse-plus-low-rank sum, jointly trained from scratch. ReDoRA and RePiSSA \citep{Shivagunde2024} extend the same idea with alternative reparameterization choices. The third category adds low-rank adapters that are periodically merged back and re-initialized. ReLoRA \citep{Lialin2023ReLoRAHT}, the canonical example, sparked the initial interest in low-rank pre-training; SwitchLoRA \citep{zhou2024switchlora} extends it by periodically switching parameters of the LoRA matrices, and LoQT \citep{Loeschcke2024LoQTLR} adapts the recipe to quantized models.

\paragraph{Toward understanding the training dynamics of low-rank methods.}
A handful of studies probe low-rank training beyond final validation loss, but each examines a single signal in isolation. Flat-LoRA~\citep{flatlora} and~\citep{plnumber} take a loss-landscape view. SLTrain~\citep{Han2024SLTrainAS} tracks singular-value spectra to argue that the sparse-plus-low-rank parameterization preserves rank that pure low-rank loses, ReLoRA~\citep{Lialin2023ReLoRAHT} inspects the rank of cumulative updates to argue that periodic merges grow effective rank beyond a single low-rank stage, and WeLore~\citep{jaiswal2024welore} characterizes the non-uniform low-rank structure of pre-trained LLM weights across layers and exploits it for fine-tuning. 


\paragraph{Geometry- and spectrum-based analyses of trained models.}
Low-rank training methods approximate full-rank training through different design choices, ranging from optimizer design to model architecture change. \citep{Li2018Visualizing} pioneers the visualization of the loss landscape itself and shows that design choices --- such as architecture, optimizer, and batch size --- shape the underlying loss landscape, and that some landscapes are far easier for an optimizer to navigate than others. Their study, however, was conducted on image classifiers, not LLMs. Subsequent work has used loss landscapes, weight spectra, and activation similarity to study different aspects of LLM training. For example, \citep{wen2024understanding} uses 1-D loss profiles to argue that LLM training under warmup-stable-decay schedules traces a ``river valley'' geometry. \citep{basin_chen2025unveiling} adapt random-direction perturbations to LLMs to characterize the basin-like structure of their loss landscape. \citep{Liu2022SamePL} uses loss-landscape to show that two language models reaching the same pre-training loss can differ on downstream tasks, with flatness predicting the gap where perplexity does not. However, \citep{andriushchenko2023modern} and \citep{kaur2023maximum} show that sharpness alone correlates poorly with generalization, motivating us to include additional signals such as rank and activation metrics, which have been used to study other aspects of model behavior like model mergeability~\citep{zhou2026demystifying}.

\section{Evaluation framework}
\label{sec:eval-framework}
\vspace{-0.3em}

We evaluate Full-rank, GaLore, Fira, CoLA, SLTrain, and ReLoRA, at 60M, 130M, and 350M parameters using 16 metrics across four dimensions.
Section~\ref{individual_metrics} introduces the individual metrics.


\subsection{Individual metrics}
\label{individual_metrics}

\subsubsection{1-D loss landscape on random directions.}
\label{sec:individualmetric_1-Dloss}

To understand the local geometry of the loss landscape around each pre-training checkpoint, we perturb the model parameters along a single randomly chosen direction and record how the validation loss varies as we move along this direction. However, since a single direction only captures a d-dimensional slice and is not necessarily representative of the surrounding geometry, we repeat the process along $100$ independently sampled random directions and record the mean and variance of the validation loss across these directions.
Following \citep{basin_chen2025unveiling} 
, we adapt the formula to our pre-training setting where the loss is the C4 \citep{raffel2020exploring} validation loss (we pre-train our models on C4). For a single random direction $\delta$:
$
  \mathcal{L}(\alpha) \;=\; \mathcal{L}_{\text{val}}\!\left(\theta + \alpha\,\delta\right),
  \qquad \delta \sim \mathcal{N}(0, I),
  \label{eq:1-D-landscape-single}
$
where $\theta$ are the checkpoint parameters, $\alpha \in [\alpha_{\min}, \alpha_{\max}]$ is the perturbation magnitude swept on a uniform grid, and $\mathcal{L}_{\text{val}}$ is the validation loss.
$\delta$ has the same dimensions as the parameter matrix it perturbs.
We average $\mathcal{L}(\alpha)$ over $100$ independently sampled directions $\delta_1, \dots, \delta_D \overset{\text{i.i.d.}}{\sim} \mathcal{N}(0, I)$ where $D = 100$:
\begin{equation}
  \bar{\mathcal{L}}(\alpha, \delta) \;=\; \frac{1}{D}\sum_{i=1}^{D} \mathcal{L}_{\text{val}}\!\left(\theta + \alpha\,\delta_i\right),
  \qquad
  \sigma^{2}(\alpha, \delta) \;=\; \frac{1}{D}\sum_{i=1}^{D}\Bigl(\mathcal{L}_{\text{val}}(\theta + \alpha\,\delta_i) - \bar{\mathcal{L}}(\alpha)\Bigr)^{2}.
  \label{eq:1-D-landscape-mean}
\end{equation}
where $\bar{\mathcal{L}}(\alpha, \delta)$ denotes the mean of the validation loss across the $D$ random directions at perturbation magnitude $\alpha$, and $\sigma^{2}(\alpha, \delta)$ denotes its variance. For convenience, we drop $\delta$ in the notation.


\paragraph{Sharpness}
The 1-D loss landscape from the previous section yields a full curve $\bar{\mathcal{L}}(\alpha, \delta)$ per checkpoint, which is qualitative in nature. To compare methods quantitatively across checkpoints, model sizes, and training steps, we reduce each curve to a single scalar capturing how flat or sharp the local basin is around the checkpoint. Intuitively, the rate at which $\bar{\mathcal{L}}$ grows as we move away from the checkpoint ($\alpha = 0$) serves as a discrete proxy for local curvature: sharp curvature produces steep loss changes in the neighborhood, while flat curvature keeps losses close to the reference value at $\alpha = 0$. Concretely, for each direction $\delta_i$, we evaluate the loss at symmetric perturbations $\pm \alpha_j$ on a grid of $N$ offsets (where $N$ is the number of positive offsets in the perturbation grid) and measure the elevation $\Delta(\delta_i, \alpha_j) = \mathcal{L}(\delta_i, \alpha_j) - \mathcal{L}(\delta_i, 0)$ relative to the checkpoint. Sharpness (S) is then the average of these elevations across all $D$ directions and both signs of $\alpha$:
\begin{equation}
S = \frac{1}{2ND} \sum_{i=0}^{D-1}  \sum_{j = 0}^{N-1} \Bigl[\Delta(\delta_i, +\alpha_{j}) + \Delta(\delta_i, -\alpha_{j})\Bigr],
\qquad
\Delta(\delta, \alpha) = \mathcal{L}(\delta, \alpha) - \mathcal{L}(\delta, 0).
\label{eq:expected_sharpness}
\end{equation}
A larger $S$ indicates a sharper basin; a smaller $S$ indicates a flatter one.

\paragraph{Direction variance}
The variance band in the 1-D loss landscape plots is a qualitative metric, and direction variance is its quantitative counterpart. We compute it by averaging the across-direction variance $\sigma^{2}(\alpha, \delta)$ from~\eqref{eq:1-D-landscape-mean} over the offsets $N$, similar to ~\eqref{eq:expected_sharpness}). See Appendix \ref{appendix:loss-landscape-metric} for equations.

\subsubsection{1-D loss landscape on top-$K$ PCA directions.}


Random Gaussian directions probe most-case geometry around a checkpoint. We therefore complement them with a principled alternative: applying SVD to the weight matrix and perturbing along its top-$k$ singular directions. Specifically, for every two-dimensional parameter tensor $W \in \mathbb{R}^{m \times n}$ in the checkpoint $\theta$, we compute the SVD, $W = U\Sigma V^{\top}$ and form the top-k perturbation direction as $\delta^{(k)} = \sigma_k(u_k \otimes v_k)$, where $u_k$ and $v_k$ are the $k$-th left and right singular vectors and $\sigma_k$ is the corresponding singular value. The resulting 1-D loss slice is $\mathcal{L}^{\text{pca}}(\alpha; k) = \mathcal{L}_{\text{val}}\!\left(\theta + \alpha\, \delta^{(k)}\right)$. 
\subsubsection{1-D interpolation}
To assess the geometric connectivity of parameter space across training, we measure the loss along a straight-line path between two checkpoints. Given two models $\theta_A$ and $\theta_B$, the interpolated model at coefficient $\beta \in [0, 1]$ is $\theta(\beta) = (1 - \beta)\theta_A + \beta\theta_B,$ and we evaluate the validation loss $\mathcal{L}(\beta) = \mathcal{L}_{\text{val}}(\theta(\beta))$ at evenly-spaced values of $\beta$. The resulting curve characterizes the loss surface geometry along the segment connecting the two models. A flat, low-lying curve indicates that both checkpoints reside in the same wide basin, whereas a curve that peaks sharply in the interior  suggests that there is a barrier between the two models and the models exist in two different basins. Following \cite{frankle2020linear}, we quantify this as \textbf{$\text{Barrier Height (BH)} = \max_{\beta \in (0,1)} \mathcal{L}(\beta) - \frac{1}{2}\Big(\mathcal{L}(0) + \mathcal{L}(1)\Big)$}. $BH$ is the excess loss at the worst intermediate point above the average of the two endpoint losses. A barrier near zero means the two models are connected through a smooth valley, and a large positive barrier means the optimizer must cross a region of high loss to travel between them. We apply this metric in two settings. First, we interpolate between consecutive checkpoints of the same method, using pairs e.g., $({\theta}_{1000}, {\theta}_{2000}), ({\theta}_{2000}, {\theta}_{3000})$ etc. throughout training. We call this \textbf{Consecutive Checkpoints Barrier Height (CCBH)}. Second, we interpolate between checkpoints of different methods at the same training step. This cross-method interpolation, whose barrier we call the \textbf{Inter-Method Barrier Height (IMBH)}, reveals whether different training methods converge to the same basin in weight space or to geometrically distinct solutions. To our knowledge, this is the first application of barrier-height analysis across different pre-training methods rather than across seeds, epochs, or fine-tuning runs of the same method. As methods such as ReLoRA and SLTrain store weights in a factored or sparse format, we first materialize all checkpoints into a common dense full-rank weight space before interpolating, ensuring that the linear path is geometrically meaningful across methods. We exclude CoLA in the cross-method interpolation comparison as CoLA has an activation function between low-rank matrices and cannot be refactored as a dense full-rank model.

\subsubsection{Rank and spectral metrics.}
Since low-rank methods are designed to approximate full-rank training, we track the spectral structure of each method's weights across the training horizon. Following~\citep{zhou2026demystifying}, we compute a family of spectral quantities on the attention and MLP weight matrices. \textbf{Effective rank} measures the entropy of the normalized singular value spectrum, capturing how broadly the matrix's energy is distributed across directions. \textbf{Stable rank}, defined via the ratio of Frobenius to operator norms, provides a noise-robust dimensionality estimate that downweights negligible singular values. \textbf{Spectral gap} measures the relative separation between the top two singular values, indicating whether a single direction clearly dominates. \textbf{Threshold rank} counts the singular values exceeding a fixed cutoff. Along with raw weights, we also apply these metrics to the parameter updates between consecutive checkpoints. We include equations for each of these metrics in Appendix~\ref{appendix_rank_spectral}.

\subsubsection{Activation-based metrics.}
To measure how closely a low-rank checkpoint's internal representations match those of the full-rank baseline, we compare their hidden states on identical inputs. We compute \textbf{Activation L2 distance}, which measures the mean per-position Euclidean distance between the two activation matrices. \textbf{Activation cosine similarity} reports the mean directional alignment of activation vectors at each position, isolating angular mismatch independent of magnitude. \textbf{Linear CKA}~\citep{kornblith2019similarity} compares the two activation matrices at the representational level and is invariant to orthogonal rotations and isotropic scaling. Equations for each is defined in Appendix~\ref{appendix:activation_metrics}.

\section{Experimental Setup}
\label{sec:setup}

\textbf{Training:} Following \citep{Zhao2024GaLoreML}, we pre-train LLaMA-style decoders at 60M, 130M, and 350M parameters on C4 with Chinchilla-optimal token budgets. We compare full-rank, GaLore, Fira, CoLA, SLTrain, and ReLoRA (without warm start), using rank 128 at 60M and 256 at larger scales, with other hyperparameters from the original papers. \textbf{Metrics:} We compute all metrics in Section~\ref{individual_metrics} on a fixed 1000-example C4 validation subset with a shared seed. \textbf{Downstream predictor:} We fit a linear regression predicting the downstream mean from validation loss and the eight sign-consistent metrics (see Table~\ref{tab:top8_features}), evaluated under leave-one-size-out (LOSO) and leave-one-method-out (LOMO) cross-validation on 90 checkpoints (5 per method × 6 methods × 3 sizes). \textbf{Downstream evaluation:} We evaluate final checkpoints zero-shot with lm-evaluation-harness~\cite{eval-harness} on 11 tasks across five categories: commonsense (HellaSwag, PIQA, COPA, SWAG), world knowledge (OpenBookQA, ARC-Easy, QA4MRE-2013), reading comprehension (ReCoRD), grammar (BLiMP), and logical reasoning (LogiQA, LogiQA2). All tasks score above random performance and show gains with model scale. We use the t5-base tokenizer, automatic batch sizing, and 1000 bootstrap iterations; task scores use standard lm-eval metrics and are reported as mean~\(\pm\) bootstrap standard error in Tables~\ref{tab:eval_60m}--\ref{tab:eval_350m} in Appendix ~\ref{appendix:downstream_eval_tables}. We use an aggregate plot for each task category across all model sizes in the Figure ~\ref{fig:downstream_eval}. All experiments were run on NVIDIA RTX 6000 Ada Generation GPUs (48 GB). Each pre-training run and metric computation in our study fits on a single GPU.

\section{Results}
\label{sec:result}

\subsection{Individual metrics results}
\label{subsec:result_individual_metric}

\begin{figure}
    \centering
    \begin{subfigure}{\linewidth}
        \centering
        \includegraphics[width=0.95\linewidth]{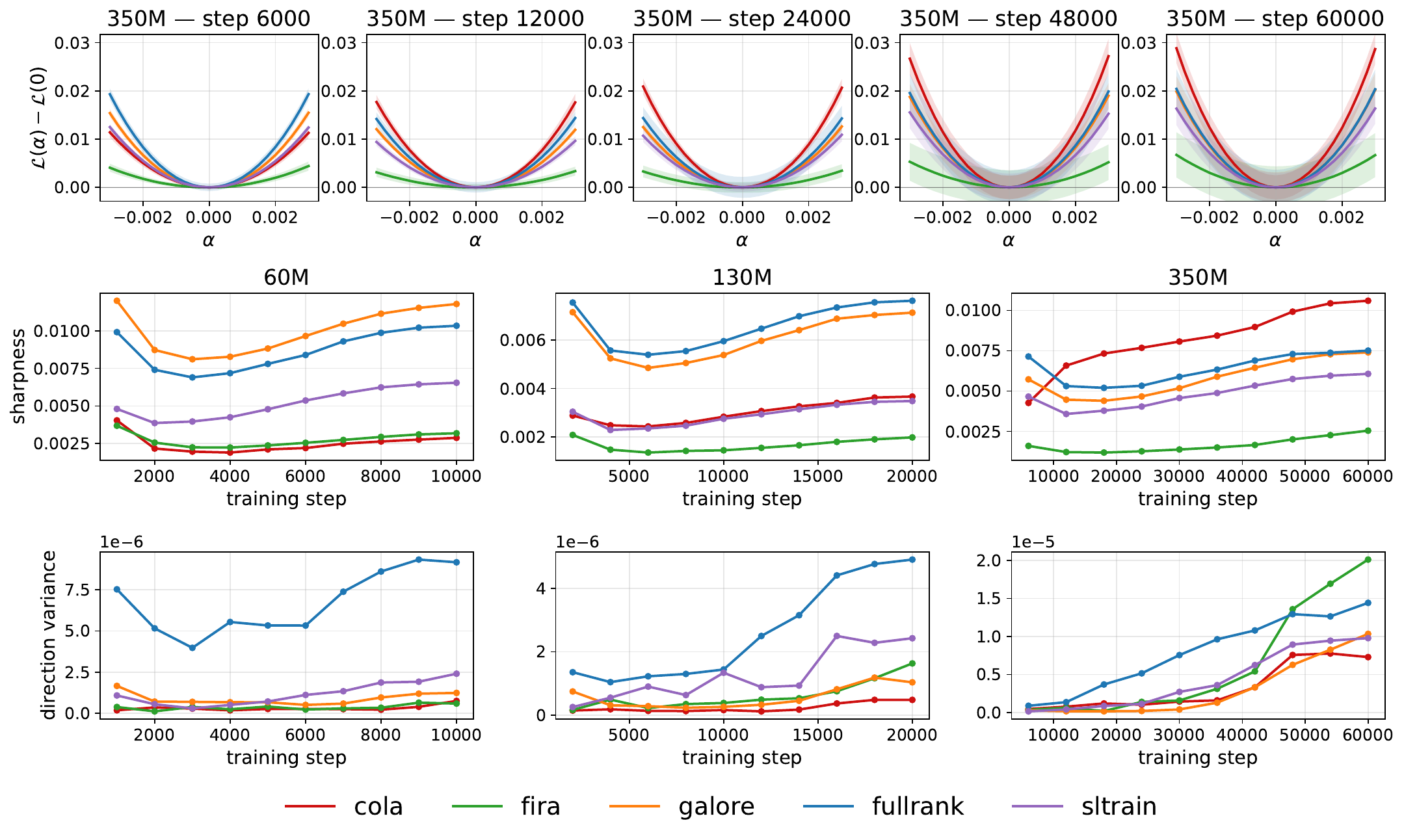}
        \caption{}
        \label{fig:loss_landscape_350m_random}
    \end{subfigure}
    
    \begin{subfigure}{\linewidth}
        \centering
        \includegraphics[width=0.9\linewidth]{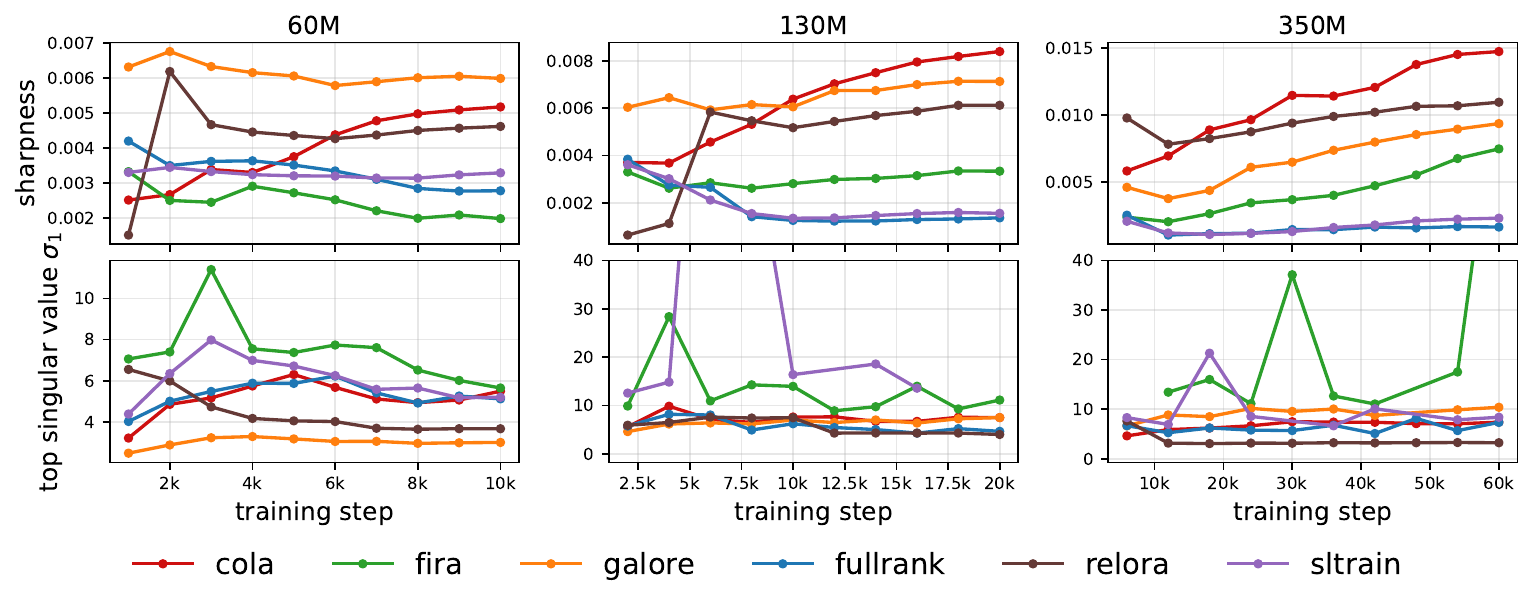}
        \caption{}
        \label{fig:loss_landscape_350m_pca-1}
    \end{subfigure}
    
    \caption{1-D loss landscape \textbf{(a)} random direction \textbf{(b)} top-1 PCA direction.}
    \label{fig:loss_landscape_pca_combined}
\end{figure}

\paragraph{1-D loss landscape on random directions}
\label{results-1-D-loss-landscape-random}
We report the 1D loss landscape along random directions in Figure~\ref{fig:loss_landscape_350m_random}: Row 1 shows the mean centered-loss curve ($\mathcal{L}(\alpha) - \mathcal{L}(0)$) with variance bands at five 350M checkpoints; Rows 2 and 3 show basin sharpness and direction variance across training for all three model sizes. 60M and 130M plots are shown in Appendix~\ref{appendix:more_results_1D_loss_landscape} (Figures~\ref{fig:landscape-60m-all-wo-relora}--\ref{fig:landscape-350m-all}).

At 60M, every method's basin \textbf{sharpness first decreases and then increases over training} --- it starts sharp at initialization, flattens during the early phase, and re-sharpens as training continues. \textbf{The sharp $\rightarrow$ flat $\rightarrow$ sharp trajectory holds at 130M and 350M (except for CoLA)}. At 350M, the basin starts sharp at initialization, flattens through the early-to-mid phase ($\sim\!12\text{k}\!-\!18\text{k}$ steps for Fira, SLTrain, GaLore, and Full-rank), and then re-sharpens as training continues. CoLA's sharpness rises monotonically from step $6\text{k}$ to $60\text{k}$, never exhibiting the intermediate flattening phase. Figure~\ref{fig:loss_landscape_350m_random} row-3 shows that  \textbf{at 60M full-rank is more anisotropic} than every low-rank method we test, except for ReLoRA. The average direction variance is of order $\sim\!10^{-5}$ for full-rank versus $10^{-7}\!-\!10^{-6}$ for low-rank pre-training methods. In other words, if we perturb a full-rank model along a random direction, the magnitude of the resulting loss change varies strongly with which direction we picked, whereas for the low-rank methods, almost any direction yields roughly the same response (loss change).
\textbf{This separation weakens at 130M}, where the low-rank curves rise toward the full-rank baseline over training. \textbf{At 350M, the gap closes more} i.e., every method reaches similar direction variance by the end of training (y-axis range is different for each model scale), and Fira actually exceeds full-rank at the final checkpoint. The clear ``full-rank-is-anisotropic, low-rank-is-uniform'' separation at 60M is therefore a small-scale artifact, as at larger scales, low-rank methods develop anisotropies comparable to or greater than full-rank training.

\paragraph{1-D loss landscape on top-1 PCA direction.} Figure~\ref{fig:loss_landscape_350m_pca-1} shows basin sharpness (top row) and top singular value $\sigma_1$ (bottom row) along the top-1 PCA direction across training. Full-rank's $\sigma_1$ is moderate ($\sim\!4\!-\!7$ across scales) and its sharpness along top-1 pca direction is low to moderate, sitting below every low-rank method except Fira 60M, suggesting full-rank basin is relatively flatter than low-rank methods. We see similar pattern with SLTrain. \textbf{Fira is the flattest along its top-1 pca direction.} Despite having one of the largest $\sigma_1$ at every scale (peaks of $\sim\!30$ at 130M and $\sim\!40$ at 350M), Fira's sharpness is the lowest which means a unit-$\alpha$ step nudges Fira further in parameter space than any other method, yet the loss barely responds, indicating its loss landscape is flattest. \textbf{GaLore, CoLA and ReLoRA converge to sharper basin than full-rank}. GaLore has relatively smaller $\sigma_1$ at every scale ($\sim\!3$ throughout training) yet still produces moderate-to-high sharpness ($\sim\!0.005\!-\!0.007$) --- the loss elevates substantially, indicating a very steep loss landscape along its leading direction. Similar pattern is seen in CoLA and ReLoRA. 
Combining random and top-1 PCA results, three regimes emerge: Fira and SLTrain converge into flat-to-moderately-flat basins; CoLA and ReLoRA reach the sharpest basins; and GaLore straddles the two — moderately flat along random directions but sharp along its top-1 PCA direction. Full-rank shows the opposite pattern to GaLore: sharp along random directions but flat along its top-1 PCA directions.

\begin{figure}
    \centering
    \begin{subfigure}{\linewidth}
        \centering
        \includegraphics[width=\linewidth]{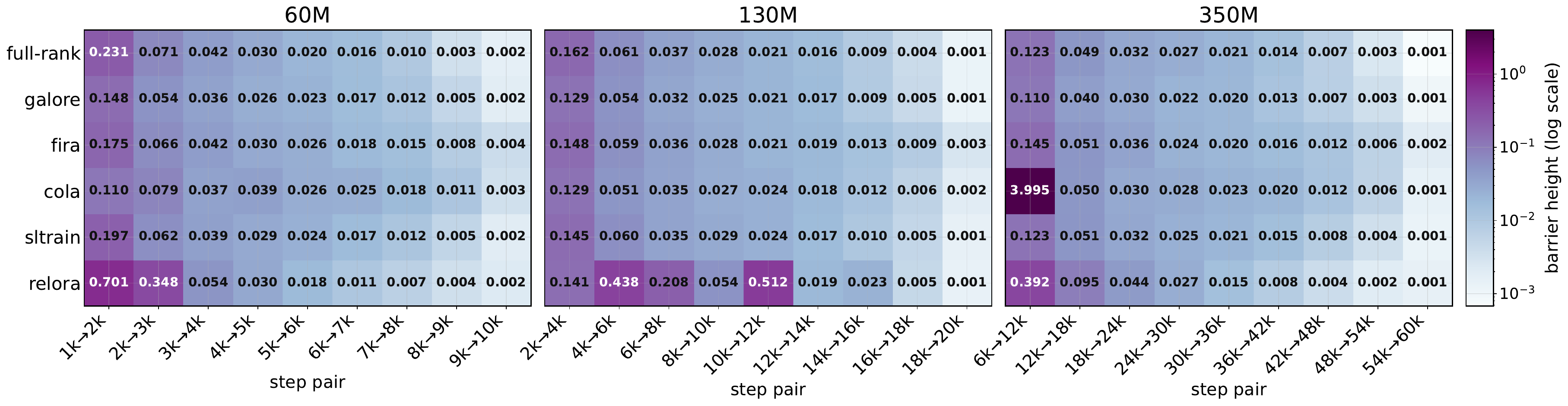}
        \caption{}
        \label{fig:interp1-D-barrier}
    \end{subfigure}
    
    \begin{subfigure}{\linewidth}
        \centering
        \includegraphics[width=\linewidth]{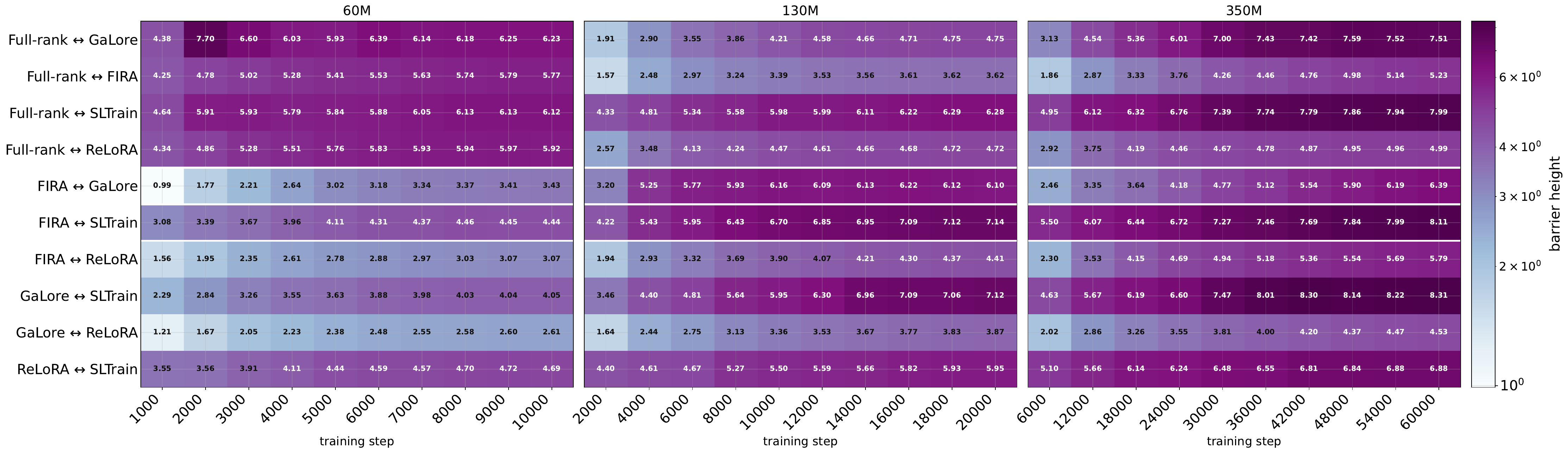}
        \caption{}
        \label{fig:crossinterp-heatmap}
    \end{subfigure}
    
    \caption{%
        1-D interpolation \textbf{(a)} CCBH \textbf{(b)} IMBH
    }
    \label{fig:interp_combined}
\end{figure}

\paragraph{1-D interpolation}
Figure \ref{fig:interp1-D-barrier} shows the CCBH heatmap for every pre-training method. We observe two patterns. First, the \textbf{CCBH decay across all methods as training progresses} (rows lighten left to right), and second, \textbf{CoLA and ReLoRA have higher CCBH than other methods}. Figure \ref{fig:crossinterp-heatmap} shows the IMBH, and we find two consistent patterns. First, every method pair sustains large barriers throughout training ($BH \approx 2$--$8$ across all sizes and steps), suggesting that all methods occupy geometrically separated basins. Second, unlike the CCBH of Figure \ref{fig:interp1-D-barrier}, which decay with training, IMBH increase monotonically. This means that methods drift further apart in weight space as each converges into its own basin. At 60M, full-rank vs.\ low-rank barriers (IMBH) are consistently higher than any low-rank vs.\ low-rank pair, indicating that the full-rank solution occupies a basin that is more distant from any low-rank method solution. Within the low-rank group at 60M, Fira, GaLore, and ReLoRA exhibit relatively low mutual barriers. SLTrain, by contrast, shows substantially higher barriers against all other low-rank methods, placing it in a distinctly separate valley. At 130M and 350M, the full-rank versus low-rank barriers decrease, and low-rank vs.\ low-rank increases. Fira and ReLoRA retain the smallest mutual barriers, GaLore occupies an intermediate position, and SLTrain remains the most geometrically isolated method at every scale. Among the low-rank methods, Fira consistently shows the lowest barrier to the full-rank solution across all model sizes (Figure \ref{fig:crossinterp-heatmap}), suggesting it converges to a basin geometrically closest to the full-rank basin. ReLoRA ranks second --- a surprise, given its highest validation loss and sharpest basin.

\begin{figure}
    \centering
    \includegraphics[width=0.9\linewidth]{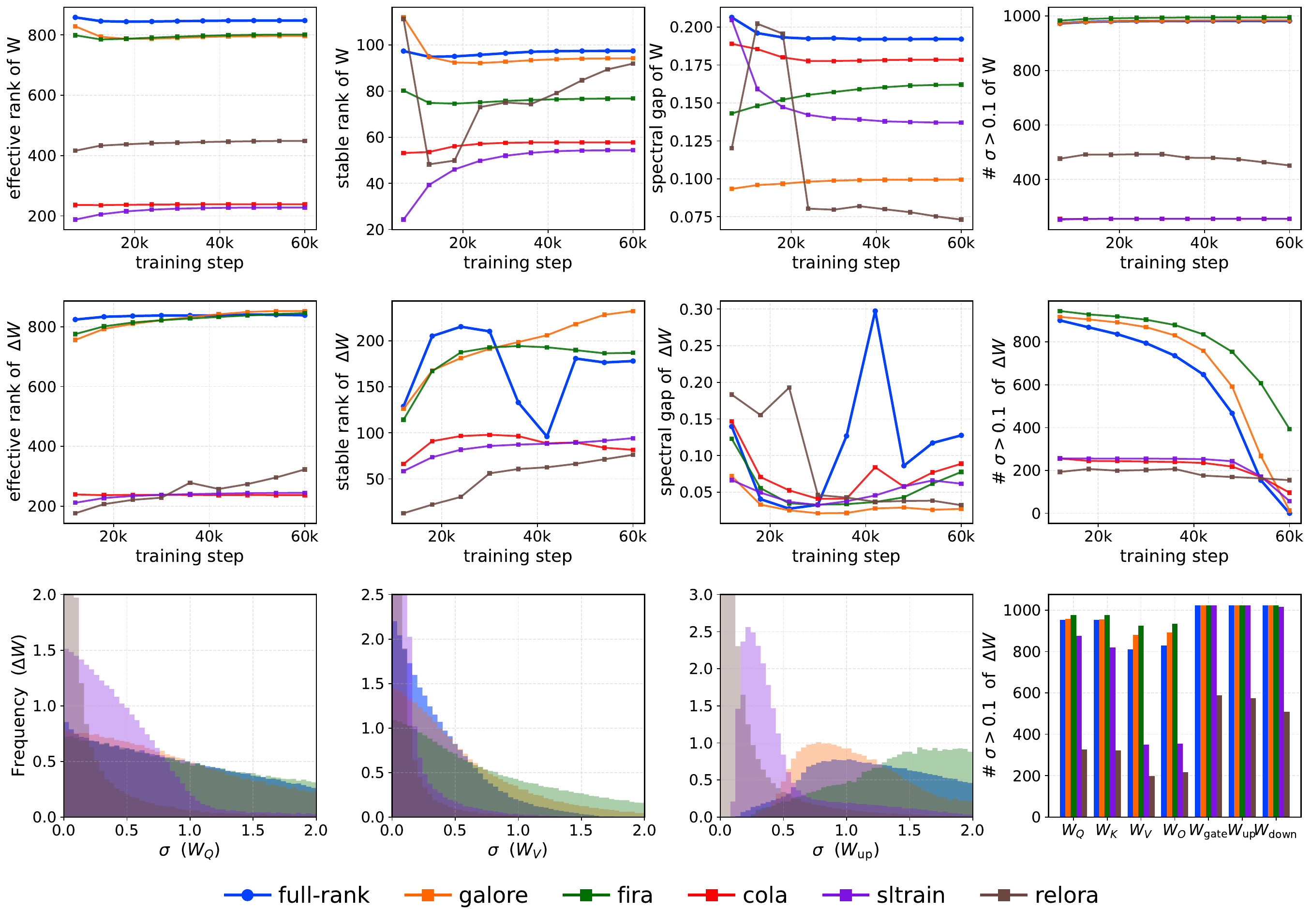}
    \caption{Rank and spectral metrics at 350M.}
    \label{fig:composite_rank}
\end{figure}

\paragraph{Rank and spectral metrics.}

Figure~\ref{fig:composite_rank} shows four rank and spectral metrics at the 350M. Row~1 tracks metrics on the trained weight $W$, Row~2 on consecutive-checkpoint updates (e.g. $\Delta W = W_{\text{2k}} - W_{\text{1k}}$), and Row~3 shows the singular-value distribution of $\Delta W$ (difference between last and first checkpoint) per projection type (attention $W_Q, W_V$ and MLP $W_{\text{up}}$), with the rightmost panel counting singular values above $0.1$ per projection. \textbf{Trained weights $W$ (Row~1).} Full-rank dominates on every rank metric and no low-rank method fully matches it. GaLore tracks full-rank closely on effective and stable rank but exhibits a substantially smaller spectral gap; CoLA shows the opposite profile. ReLoRA maintains higher effective and stable rank than both CoLA and SLTrain despite its low-rank constraint. \textbf{Weight update $\Delta W$ (Row~2).} Fira and GaLore most closely follow full-rank on effective and stable rank, while full-rank retains the largest spectral gap that no low-rank method reaches. GaLore further tracks full-rank's count of singular values above $0.1$ more faithfully than the other methods.  \textbf{Per-projection spectra (Row~3).} Within each projection type, GaLore reproduces full-rank's singular-value distribution more closely than competing methods. We see similar pattern at 60M and 130M scale (see plots in Appendix section ~\ref{appendix:results_rank}.)

\paragraph{Activation metrics.}
\label{sec:activation_alignment}
Figure~\ref{fig:composite_activation} summarizes activation alignment with full-rank training. Row~1 reports the layer-averaged stacked deviation per method and scale; Row~2 restricts the same quantity to the final layer; Row~3 shows per-layer $\times$ per-step heatmaps for the two extreme low-rank methods at 350M (Fira and CoLA). Full per-method, per-scale heatmaps are in Appendix~\ref{appendix:activation_metrics}. \textbf{Layer-averaged drift (Row~1).} ReLoRA tracks full-rank representations most closely across all three scales, with GaLore and SLTrain following, while CoLA and Fira exhibit the largest total deviation. The ranking is stable across scales. \textbf{Last-layer drift (Row~2).} Restricting to the final layer changes the picture for Fira specifically: its last-layer deviation drops sharply ($2.49$ at 350M, second lowest after ReLoRA) despite having the largest layer-averaged drift. The last decoder block plus final RMSNorm reduces the deviation, and Fira benefits from this more than any other method. \textbf{Per-layer dynamics (Row~3).} For both Fira and CoLA, $L_2$ distance grows in the later layers as training progresses, with the final layer reducing drift relative to full-rank. CoLA is directionally off at every layer ($\cos \approx 0$), while Fira preserves angular alignment. Linear CKA deviates most in the middle layers and degrades further later in training. Patterns hold across other methods and scales (Figures~\ref{fig:act-last-l2-layers}--\ref{fig:act-cosine-layers}, Appendix~\ref{appendix:results_activation_full}).

\vspace{-0.3em}
\subsection{Downstream evaluation}
\label{subsec:val_vs_downstream}
\begin{figure}
\centering
\begin{minipage}[c]{0.30\linewidth}
\centering
\small
\setlength{\tabcolsep}{4pt}
\begin{tabular}{lccc}
\toprule
Method & 60M & 130M & 350M \\
\midrule
Full-rank        & 34.12             & \underline{25.09} & \underline{19.43} \\
\textsc{GaLore}  & 34.91             & 25.62             & 19.87 \\
\textsc{Fira}    & \textbf{31.13}    & \textbf{23.18}    & \textbf{17.40} \\
\textsc{CoLA}    & \underline{33.47} & 25.79             & 19.58 \\
\textsc{SLTrain} & 35.71             & 26.48             & 20.64 \\
\textsc{ReLoRA}  & 42.76             & 35.88             & 29.99 \\
\bottomrule
\end{tabular}
\captionof{table}{\small Validation perplexity at last checkpoint, across model sizes. Lower is better. Per column, the best method is \textbf{bold} and second-best is \underline{underlined}.}
\label{tab:eval_perplexity_by_size}
\end{minipage}%
\hspace{0.03\linewidth}%
\begin{minipage}[c]{0.55\linewidth}
\centering
\includegraphics[width=0.85\linewidth]{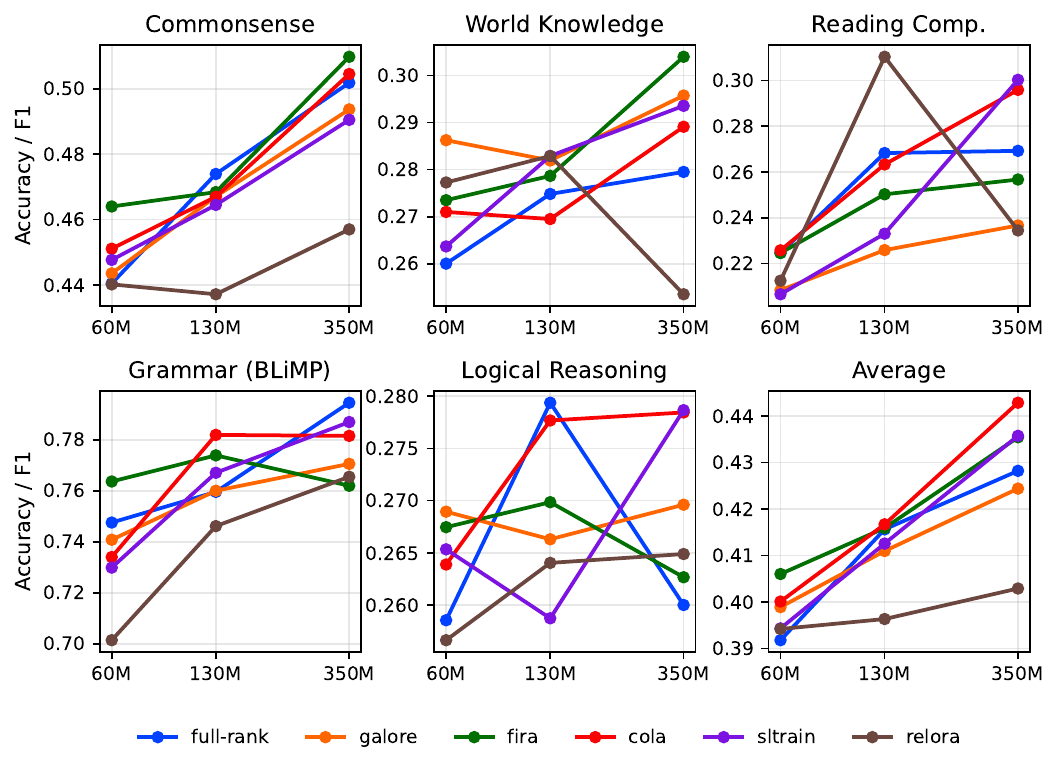}
\captionof{figure}{\small Zero-shot downstream performance.}
\label{fig:downstream_eval}
\end{minipage}

\begin{minipage}[c]{0.88\linewidth}
\centering
\small
\setlength{\tabcolsep}{6pt}
\begin{tabular}{lccc}
\toprule
Predictor & LOSO Pearson & LOMO Pearson & $R^2$ (in-sample) \\
\midrule
val loss only & 0.873 & 0.864 & 0.841 \\
geometry only (8 feats) & 0.498 & 0.431 & 0.558 \\
val loss + geometry (9 feats) & \textbf{0.913} & \textbf{0.895} & \textbf{0.907} \\
\bottomrule
\end{tabular}
\captionof{table}{\small Comparison of downstream predictors.  Geometry features are the top-8 paper metrics whose per-size Spearman vs downstream is sign-consistent (Spearman correlation with downstream has the same sign at all three scales). \textbf{Bold} = best per column.}
\label{tab:predictor_comparison}
\end{minipage}
\end{figure}

Table~\ref{tab:eval_perplexity_by_size} reports final C4 validation perplexity and Figure~\ref{fig:downstream_eval} reports zero-shot downstream performance. \textbf{Validation perplexity does not predict downstream ranking.} Fira achieves the lowest perplexity at every scale, even surpassing full-rank, yet CoLA leads on multiple downstream categories at 130M and 350M despite not ranking in the top two on perplexity. \textbf{Geometry and spectral metrics recover what perplexity misses.} Across 90 checkpoints, perplexity alone is a strong but incomplete predictor (LOSO Pearson 0.873, LOMO 0.864); augmenting it with the eight sign-consistent geometry and spectral features in Table~\ref{tab:top8_features} lifts LOSO to 0.913 and LOMO to 0.895 (Table~\ref{tab:predictor_comparison}).

\vspace{-0.3em}
\section{Conclusion}
\vspace{-0.5em}
Low-rank pre-training methods are not interchangeable approximations of full-rank training. Fira and SLTrain reach flat basins while CoLA and ReLoRA reach the sharpest; each method converges to a geometrically distinct basin, with inter-method barriers growing rather than shrinking over training; GaLore most faithfully reproduces full-rank's spectral structure on weight updates, while ReLoRA tracks full-rank activations most closely. Validation perplexity shows an incomplete picture of downstream performance — but augmenting it with eight geometric and spectral features from our framework recovers the missing signal, providing a better predictor of downstream performance.

\begin{figure}
    \centering
    \includegraphics[width=0.95\linewidth]{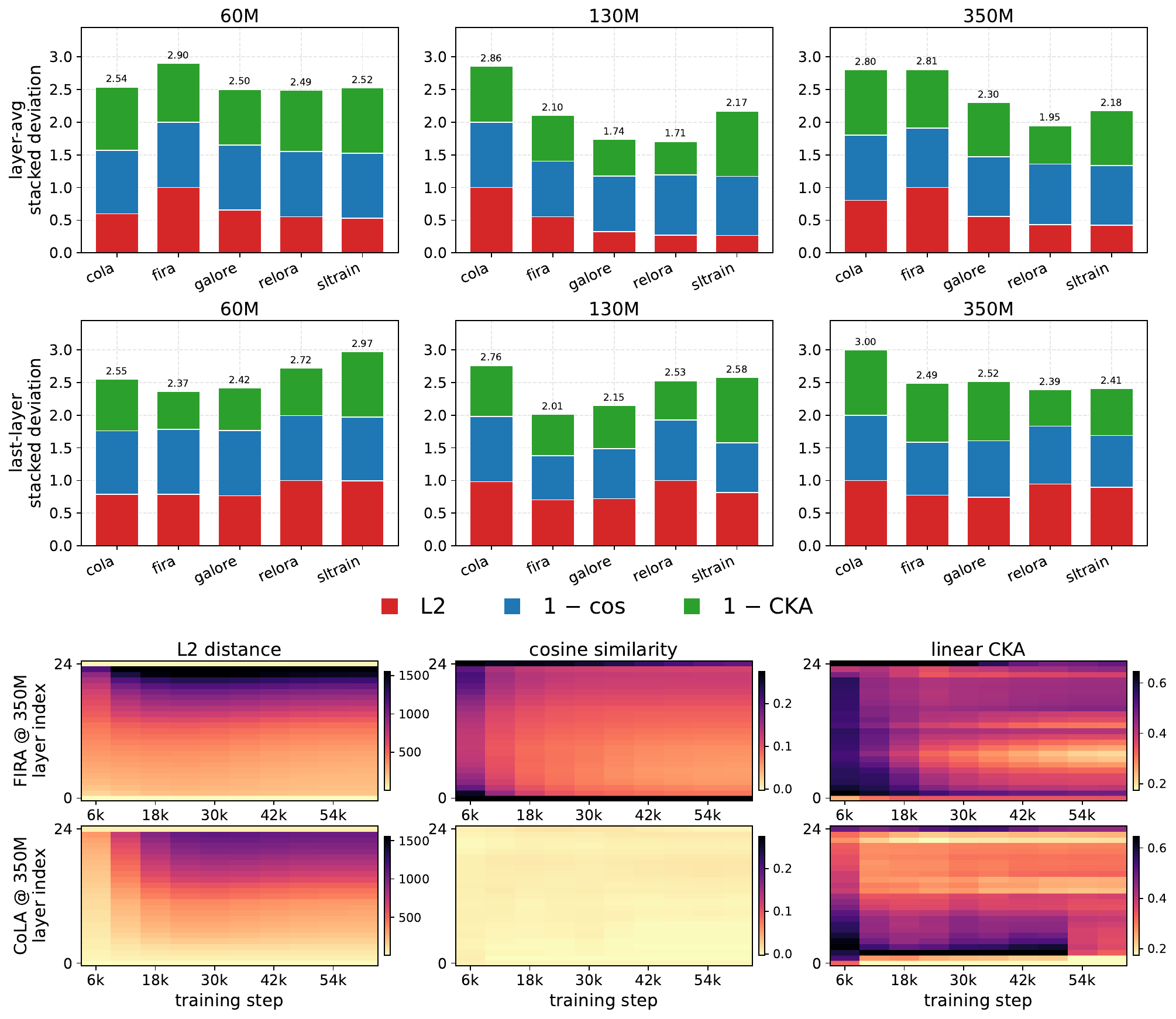}
    \caption{Activation deviation with full-rank baseline.}
    \label{fig:composite_activation}
\end{figure}

\vspace{-0.5em}
\section{Limitations}
\label{sec:limitations}
We compute our metrics on $<1B$ LLaMA-style models, generalization to billion-scale models, other architectures, datasets, and rank choices is not explored. Due to compute limit, we report each metric from a single training run per method per scale and therefore cannot quantify run-to-run variability. We adopt two perturbation-direction choices for the 1-D loss landscape, but the most appropriate normalization for cross-method LLMs comparison remains open.

\bibliographystyle{plain}
\bibliography{references}

@article{llama,
  title={LLaMA: Open and Efficient Foundation Language Models},
  author={Touvron, Hugo and Lavril, Thibaut and Izacard, Gautier and Martinet, Xavier and Lachaux, Marie-Anne and Lacroix, Timoth{\'e}e and Rozi{\`e}re, Baptiste and Goyal, Naman and Hambro, Eric and Azhar, Faisal and Rodriguez, Aurelien and Joulin, Armand and Grave, Edouard and Lample, Guillaume},
  journal={arXiv preprint arXiv:2302.13971},
  year={2023}
}

@inproceedings{Lialin2023ReLoRAHT,
  author       = {Vladislav Lialin and
                  Sherin Muckatira and
                  Namrata Shivagunde and
                  Anna Rumshisky},
  title        = {{R}e{L}o{RA}: {H}igh-{R}ank {T}raining {T}hrough {L}ow-{R}ank {U}pdates},
  booktitle    = {The Twelfth International Conference on Learning Representations,
                  {ICLR} 2024, Vienna, Austria, May 7-11, 2024},
  publisher    = {OpenReview.net},
  year         = {2024},
  url          = {https://openreview.net/forum?id=DLJznSp6X3},
  bibsource    = {dblp computer science bibliography, https://dblp.org}
}

@inproceedings{Zhao2024GaLoreML,
  author       = {Jiawei Zhao and
                  Zhenyu Zhang and
                  Beidi Chen and
                  Zhangyang Wang and
                  Anima Anandkumar and
                  Yuandong Tian},
  title        = {Ga{L}ore: {M}emory-{E}fficient {LLM} {T}raining by {G}radient {L}ow-{R}ank {P}rojection},
  booktitle    = {Forty-first International Conference on Machine Learning, {ICML} 2024,
                  Vienna, Austria, July 21-27, 2024},
  publisher    = {OpenReview.net},
  year         = {2024},
  url          = {https://openreview.net/forum?id=hYHsrKDiX7},
  bibsource    = {dblp computer science bibliography, https://dblp.org}
}

@article{Han2024SLTrainAS,
  title={{SLT}rain: a sparse plus low-rank approach for parameter and memory efficient pretraining},
  author={Han, Andi and Li, Jiaxiang and Huang, Wei and Hong, Mingyi and Takeda, Akiko and Jawanpuria, Pratik and Mishra, Bamdev},
  journal={arXiv preprint arXiv:2406.02214},
  year={2024}
}

@article{Loeschcke2024LoQTLR,
  title={{L}o{QT}: {L}ow {R}ank {A}dapters for {Q}uantized {T}raining},
  author={Loeschcke, Sebastian and Toftrup, Mads and Kastoryano, Michael J and Belongie, Serge and Sn{\ae}bjarnarson, V{\'e}steinn},
  journal={arXiv preprint arXiv:2405.16528},
  year={2024}
}

@article{Muhamed2024GrassCE,
  title={Grass: {C}ompute efficient low-memory llm training with structured sparse gradients},
  author={Muhamed, Aashiq and Li, Oscar and Woodruff, David and Diab, Mona and Smith, Virginia},
  journal={arXiv preprint arXiv:2406.17660},
  year={2024}
}

@article{Chen2024FiraCW,
  title={Fira: Can We Achieve Full-rank Training of LLMs Under Low-rank Constraint?},
  author={Xi Chen and Kaituo Feng and Changsheng Li and Xunhao Lai and Xiangyu Yue and Ye Yuan and Guoren Wang},
  journal={ArXiv},
  year={2024},
  volume={abs/2410.01623},
  url={https://api.semanticscholar.org/CorpusID:273026172}
}

@inproceedings{liu2025cola,
  title={Cola: Compute-efficient pre-training of llms via low-rank activation},
  author={Liu, Ziyue and Zhang, Ruijie and Wang, Zhengyang and Yan, Mingsong and Yang, Zi and Hovland, Paul D and Nicolae, Bogdan and Cappello, Franck and Tang, Sui and Zhang, Zheng},
  booktitle={Proceedings of the 2025 Conference on Empirical Methods in Natural Language Processing},
  pages={4627--4645},
  year={2025}
}

@inproceedings{plnumber,
  title = {On the Optimization Landscape of Low Rank Adaptation Methods for Large Language Models},
  author = {Liu, Xu-Hui and Du, Yali and Wang, Jun and Yu, Yang},
  booktitle = {The Thirteenth International Conference on Learning Representations},
  year = {2025},
}

@article{flatlora,
  title={Flat-lora: Low-rank adaptation over a flat loss landscape},
  author={Li, Tao and He, Zhengbao and Li, Yujun and Wang, Yasheng and Shang, Lifeng and Huang, Xiaolin},
  journal={arXiv preprint arXiv:2409.14396},
  year={2024}
}

@article{basin_chen2025unveiling,
  title={Unveiling the Basin-Like Loss Landscape in Large Language Models},
  author={Chen, Huanran and Dong, Yinpeng and Wei, Zeming and Huang, Yao and Zhang, Yichi and Su, Hang and Zhu, Jun},
  journal={arXiv preprint arXiv:2505.17646},
  year={2025}
}

@article{zhou2026demystifying,
  title   = {Demystifying Mergeability: Interpretable Properties to Predict Model Merging Success},
  author  = {Zhou, Luca and Zhao, Bo and Yu, Rose and Rodol{\`a}, Emanuele},
  journal = {arXiv preprint arXiv:2601.22285},
  year    = {2026},
  url     = {https://arxiv.org/abs/2601.22285},
}

@inproceedings{kornblith2019similarity,
  title     = {Similarity of Neural Network Representations Revisited},
  author    = {Kornblith, Simon and Norouzi, Mohammad and Lee, Honglak and Hinton, Geoffrey},
  booktitle = {International Conference on Machine Learning (ICML)},
  year      = {2019},
  url       = {https://arxiv.org/abs/1905.00414},
}

@inproceedings{frankle2020linear,
  title={Linear mode connectivity and the lottery ticket hypothesis},
  author={Frankle, Jonathan and Dziugaite, Gintare Karolina and Roy, Daniel and Carbin, Michael},
  booktitle={International conference on machine learning},
  pages={3259--3269},
  year={2020},
  organization={PMLR}
}

@inproceedings{Liu2022SamePL,
  title={Same Pre-training Loss, Better Downstream: Implicit Bias Matters for Language Models},
  author={Hong Liu and Sang Michael Xie and Zhiyuan Li and Tengyu Ma},
  booktitle={International Conference on Machine Learning},
  year={2022},
  url={https://api.semanticscholar.org/CorpusID:253107233}
}

@article{miles2024velora,
  title={Velora: Memory efficient training using rank-1 sub-token projections},
  author={Miles, Roy and Reddy, Pradyumna and Elezi, Ismail and Deng, Jiankang},
  journal={Advances in Neural Information Processing Systems},
  volume={37},
  pages={42292--42310},
  year={2024}
}

@article{li2025lost,
  title={Lost: Low-rank and sparse pre-training for large language models},
  author={Li, Jiaxi and Yin, Lu and Shen, Li and Xu, Jinjin and Xu, Liwu and Huang, Tianjin and Wang, Wenwu and Liu, Shiwei and Wang, Xilu},
  journal={arXiv preprint arXiv:2508.02668},
  year={2025}
}

@article{zhou2024switchlora,
  title={SwitchLoRA: Switched Low-Rank Adaptation Can Learn Full-Rank Information},
  author={Zhou, Kaiye and Wang, Shucheng and Xu, Jun},
  journal={arXiv preprint arXiv:2406.06564},
  year={2024}
}

@article{zhang2024q,
  title={Q-galore: Quantized galore with int4 projection and layer-adaptive low-rank gradients},
  author={Zhang, Zhenyu and Jaiswal, Ajay and Yin, Lu and Liu, Shiwei and Zhao, Jiawei and Tian, Yuandong and Wang, Zhangyang},
  journal={arXiv preprint arXiv:2407.08296},
  year={2024}
}

@article{Su2025GaLore2L,
  title={GaLore 2: Large-Scale LLM Pre-Training by Gradient Low-Rank Projection},
  author={DiJia Su and Andrew Gu and Jane Xu and Yuan Tian and Jiawei Zhao},
  journal={ArXiv},
  year={2025},
  volume={abs/2504.20437},
  url={https://api.semanticscholar.org/CorpusID:278171328}
}

@inproceedings{huang2024galore,
  title={Galore-mini: Low rank gradient learning with fewer learning rates},
  author={Huang, Weihao and Zhang, Zhenyu and Zhang, Yushun and Luo, Zhi-Quan and Sun, Ruoyu and Wang, Zhangyang},
  booktitle={NeurIPS 2024 Workshop on Fine-Tuning in Modern Machine Learning: Principles and Scalability},
  year={2024}
}

@article{hao2024flora,
  title={Flora: Low-rank adapters are secretly gradient compressors},
  author={Hao, Yongchang and Cao, Yanshuai and Mou, Lili},
  journal={arXiv preprint arXiv:2402.03293},
  year={2024}
}

@inproceedings{xiao2025coap,
  title={Coap: Memory-efficient training with correlation-aware gradient projection},
  author={Xiao, Jinqi and Sang, Shen and Zhi, Tiancheng and Liu, Jing and Yan, Qing and Luo, Linjie and Yuan, Bo},
  booktitle={Proceedings of the IEEE/CVF Conference on Computer Vision and Pattern Recognition},
  pages={30116--30126},
  year={2025}
}

@Article{Shivagunde2024,
 author = {Namrata Shivagunde and Mayank Kulkarni and Giannis Karamanolakis and Jack G. M. FitzGerald and Yannick Versley and Saleh Soltan and Volkan Cevher and Jianhua Lu and Anna Rumshisky},
 title = {Approximations may be all you need: Towards pre-training LLMs with low-rank decomposition and optimizers},
 year = {2024},
 url = {https://www.amazon.science/publications/approximations-may-be-all-you-need-towards-pre-training-llms-with-low-rank-decomposition-and-optimizers},
}

@article{jaiswal2024welore,
  title={From galore to welore: How low-rank weights non-uniformly emerge from low-rank gradients},
  author={JAISWAL, AJAY KUMAR and Yin, Lu and Zhang, Zhenyu and Liu, Shiwei and Zhao, Jiawei and Tian, Yuandong and Wang, Zhangyang},
  url = {https://arxiv.org/html/2407.11239v1#:~:text=In%20this%20section%2C%20we%20investigate,GaLore%20in%20different%20compression%20ratios.}
}

@inproceedings{Li2018Visualizing,
  author    = {Hao Li and Zheng Xu and Gavin Taylor and Christoph Studer and Tom Goldstein},
  title     = {Visualizing the Loss Landscape of Neural Nets},
  booktitle = {Advances in Neural Information Processing Systems (NeurIPS)},
  year      = {2018},
  url       = {https://arxiv.org/abs/1712.09913}
}

@article{wen2024understanding,
  title={Understanding warmup-stable-decay learning rates: A river valley loss landscape perspective},
  author={Wen, Kaiyue and Li, Zhiyuan and Wang, Jason and Hall, David and Liang, Percy and Ma, Tengyu},
  journal={arXiv preprint arXiv:2410.05192},
  year={2024}
}

@misc{eval-harness,
  author       = {Gao, Leo and Tow, Jonathan and Abbasi, Baber and Biderman, Stella and Black, Sid and DiPofi, Anthony and Foster, Charles and Golding, Laurence and Hsu, Jeffrey and Le Noac'h, Alain and Li, Haonan and McDonell, Kyle and Muennighoff, Niklas and Ociepa, Chris and Phang, Jason and Reynolds, Laria and Schoelkopf, Hailey and Skowron, Aviya and Sutawika, Lintang and Tang, Eric and Thite, Anish and Wang, Ben and Wang, Kevin and Zou, Andy},
  title        = {The Language Model Evaluation Harness},
  month        = 07,
  year         = 2024,
  publisher    = {Zenodo},
  version      = {v0.4.3},
  doi          = {10.5281/zenodo.12608602},
  url          = {https://zenodo.org/records/12608602}
}

@article{andriushchenko2023modern,
  title={A modern look at the relationship between sharpness and generalization},
  author={Andriushchenko, Maksym and Croce, Francesco and M{\"u}ller, Maximilian and Hein, Matthias and Flammarion, Nicolas},
  journal={arXiv preprint arXiv:2302.07011},
  year={2023}
}

@InProceedings{kaur2023maximum,
  title = 	 {On the Maximum Hessian Eigenvalue and Generalization},
  author =       {Kaur, Simran and Cohen, Jeremy and Lipton, Zachary Chase},
  booktitle = 	 {Proceedings on "I Can't Believe It's Not Better!  - Understanding Deep Learning Through Empirical Falsification" at NeurIPS 2022 Workshops},
  pages = 	 {51--65},
  year = 	 {2023},
  editor = 	 {Antorán, Javier and Blaas, Arno and Feng, Fan and Ghalebikesabi, Sahra and Mason, Ian and Pradier, Melanie F. and Rohde, David and Ruiz, Francisco J. R. and Schein, Aaron},
  volume = 	 {187},
  series = 	 {Proceedings of Machine Learning Research},
  month = 	 {03 Dec},
  publisher =    {PMLR},
  url = 	 {https://proceedings.mlr.press/v187/kaur23a.html}
}

@article{raffel2020exploring,
  title={Exploring the limits of transfer learning with a unified text-to-text transformer},
  author={Raffel, Colin and Shazeer, Noam and Roberts, Adam and Lee, Katherine and Narang, Sharan and Matena, Michael and Zhou, Yanqi and Li, Wei and Liu, Peter J},
  journal={Journal of machine learning research},
  volume={21},
  number={140},
  pages={1--67},
  year={2020}
}

\medskip

\small

\appendix

\section{More details on metrics}
\label{appendix:rank}

We provide more details on the metrics in this section.

\subsection{Loss landscape related metrics}
\label{appendix:loss-landscape-metric}
Direction variance equation is given below
$
  DV \;=\; \frac{1}{2N}\sum_{j=1}^{N} \Bigl[\sigma^{2}(+\alpha_{j}) + \sigma^{2}(-\alpha_{j})\Bigr],
  \label{eq:direction-variance}
$
A higher $DV$ indicates that the loss surface is highly anisotropic around the checkpoint, i.e., the basin is much steeper along some directions than others, whereas a lower $DV$ indicates a more isotropic basin where the loss grows at a similar rate regardless of the sampled direction.

\subsection{Rank related metrics}
\label{appendix_rank_spectral}
Since each low-rank pre-training method imposes a different structural prior on the weight matrices it learns, we examine the singular value distribution of the trained checkpoints to characterize how that prior manifests across the training horizon. These metrics surface complementary aspects of the spectrum: how much of the matrix's energy concentrates in the leading components, how broadly the remaining mass is distributed, and how many directions actively contribute to the matrix's action. Following the probe family adopted by~\citep{zhou2026demystifying} in their study of task-vector alignment, we restrict our analysis to the attention and MLP weight matrices, as these are the layers where the low-rank parametrization is enforced. For each two-dimensional weight $W \in \mathbb{R}^{m \times n}$, we obtain the ordered singular values $\sigma_{1} \geq \sigma_{2} \geq \dots \geq \sigma_{r} \geq 0$ with $r = \min(m, n)$ and compute the following quantities.

\begin{itemize}
  \item \textbf{Effective rank:} The exponential of the entropy of the normalized spectrum, treating singular values as a probability distribution:
  \begin{equation}
    \mathrm{EffRank}(W) \;=\; \exp\!\left(-\sum_{i=1}^{r} p_{i}\log p_{i}\right),
    \qquad p_{i} \;=\; \frac{\sigma_{i}}{\sum_{j} \sigma_{j}}.
    \label{eq:effective-rank}
  \end{equation}
  This quantity equals $1$ when the spectrum collapses entirely onto a single direction and approaches $r$ as the singular values flatten toward uniformity.
  \item \textbf{Stable rank:} A spectrum-based dimensionality measure defined as the ratio of the Frobenius norm to the operator norm,
  \begin{equation}
    \mathrm{StableRank}(W) \;=\; \frac{\|W\|_{F}^{2}}{\|W\|_{2}^{2}} \;=\; \frac{\sum_{i} \sigma_{i}^{2}}{\sigma_{1}^{2}}.
    \label{eq:stable-rank}
  \end{equation}
  It is upper-bounded by the algebraic rank but, unlike the latter, is insensitive to negligibly small singular values. A value of $1$ corresponds to an essentially rank-one matrix, while a value approaching $r$ reflects a near-uniform spectrum.
  \item \textbf{Spectral gap:} The relative separation between the leading two singular values, $\mathrm{SpectralGap}(W) = (\sigma_{1} - \sigma_{2})/\sigma_{1}$. A pronounced gap signals that the top direction stands out clearly from the rest of the spectrum; a vanishing gap means the first two directions carry comparable weight and are not robustly distinguishable.
  \item \textbf{Threshold rank:} A hard count of singular values whose magnitude exceeds a fixed cutoff $\tau = 0.1$, namely $\mathrm{TRank}_{\tau}(W) = \#\{\,i : \sigma_{i} > \tau\,\}$. This filters out tail components that are numerically nonzero but have negligible influence on the matrix's action, providing a noise-robust alternative to the strict rank.
\end{itemize}

\subsection{Activation-based metrics.}
\label{appendix:activation_metrics}
To measure how closely a low-rank checkpoint's internal representations match those of the full-rank baseline, we compare their hidden states on identical inputs. Using the same fixed evaluation subset and seed as in section \ref{sec:individualmetric_1-Dloss}, we run both the full-rank baseline $f_{A}$ and the low-rank target $f_{B}$ and extract the hidden states at every layer $\ell \in \{0, 1, \dots, L\}$, where $\ell=0$ denotes the embedding output. The resulting per-layer activation matrices are
$$
H^{(\ell)}_{A} \in \mathbb{R}^{N \times d}, \qquad
H^{(\ell)}_{B} \in \mathbb{R}^{N \times d},
$$
respectively, where $N$ is the total number of token positions in a sequence and $d$ is the hidden dimension. Let $h^{(\ell)}_{A,i}$ and $h^{(\ell)}_{B,i}$ denote the $i$-th row of $H^{(\ell)}_{A}$ and $H^{(\ell)}_{B}$ respectively, i.e.\ the hidden activation vectors at position $i$. We then report the following per-layer metrics.

\begin{itemize}
  \item \textbf{Activation L2 distance:} the mean per-position Euclidean distance between the two activation matrices,
  \begin{equation}
    d_{\text{L2}}^{(\ell)} \;=\; \frac{1}{N}\sum_{i=1}^{N}\left\|h^{(\ell)}_{A,i} - h^{(\ell)}_{B,i}\right\|_{2}.
    \label{eq:act-l2}
  \end{equation}
  A value of $0$ means the two checkpoints produce identical activations on the input, and larger values indicate that the low-rank model departs further from the baseline in the absolute feature space.
  \item \textbf{Activation cosine similarity:} the mean per-position directional alignment between the two activation vectors,
  \begin{equation}
    \mathrm{cos}^{(\ell)} \;=\; \frac{1}{N}\sum_{i=1}^{N}\frac{\langle h^{(\ell)}_{A,i},\, h^{(\ell)}_{B,i}\rangle}{\|h^{(\ell)}_{A,i}\|_{2}\,\|h^{(\ell)}_{B,i}\|_{2} + \varepsilon}.
    \label{eq:act-cos}
  \end{equation}
  A value of $1$ indicates that, at every position, the low-rank model points in the same direction in feature space as the baseline; values near $0$ indicate orthogonal, unrelated features. Unlike the L2 distance, this is insensitive to activation magnitude and isolates the directional component of the mismatch.
    \item \textbf{Linear CKA:} following~\citep{kornblith2019similarity}, we compute the linear Centered Kernel Alignment between the two full activation matrices as a representation-level similarity. Let $\tilde H^{(\ell)}_{A}$ and $\tilde H^{(\ell)}_{B}$ denote the column-centered versions of $H^{(\ell)}_{A}$ and $H^{(\ell)}_{B}$, and define the cross- and self-Gram matrices
  $$
    G_{AB} \;=\; \bigl(\tilde H^{(\ell)}_{B}\bigr)^{\!\top} \tilde H^{(\ell)}_{A},
    \qquad
    G_{AA} \;=\; \bigl(\tilde H^{(\ell)}_{A}\bigr)^{\!\top} \tilde H^{(\ell)}_{A},
    \qquad
    G_{BB} \;=\; \bigl(\tilde H^{(\ell)}_{B}\bigr)^{\!\top} \tilde H^{(\ell)}_{B}.
  $$
  Linear CKA is then
  \begin{equation}
    \mathrm{CKA}^{(\ell)} \;=\; \frac{\left\| G_{AB} \right\|_{F}^{2}}{\left\| G_{AA} \right\|_{F}\,\left\| G_{BB} \right\|_{F}}.
    \label{eq:act-cka}
  \end{equation} 
  Linear CKA is invariant to orthogonal rotations and isotropic scaling of the hidden features, so it captures whether the two checkpoints encode the same representational subspace even when individual neurons have been permuted or re-scaled. A value of $1$ indicates perfectly aligned subspaces and a value near $0$ indicates unrelated ones.
\end{itemize}

\section{Downstream evaluation}
\label{appendix:downstream_eval_tables}

We report downstream performance per task for 60M, 130M, 350M in Table ~\ref{tab:eval_60m}, ~\ref{tab:eval_130m}, and ~\ref{tab:eval_350m}, respectively.

\begin{table*}[h]
\centering
\small
\setlength{\tabcolsep}{4pt}
\caption{60M models final checkpoint evaluation on downstream tasks. Per-task results as presented as mean $\pm$ bootstrap standard error. Accuracy/F1 metrics are shown in \%. Section abbreviations: CS = Commonsense, WK = World Knowledge, RC = Reading Comprehension, Gr.\ = Grammar, LR = Logical Reasoning, IT = Instruction / Truthfulness. Metric symbols: $\dagger$ length-normalised accuracy, $\ddagger$ F1, $\downarrow$ lower-is-better. Per task, the best method is \textbf{bold} and the second-best is \underline{underlined}; \textit{n/a} marks tasks for which bootstrap stderr is unavailable. The Random column gives the chance-level accuracy of each task in \%; ``--'' marks tasks without a well-defined chance baseline (F1 / perplexity).}
\label{tab:eval_60m}
\resizebox{\textwidth}{!}{%
\begin{tabular}{lccccccc}
\toprule
Task & Random & \textsc{LLaMA} & \textsc{GaLore} & \textsc{Fira} & \textsc{CoLA} & \textsc{SLTrain} & \textsc{ReLoRA} \\
\midrule
\multicolumn{8}{l}{\textit{CS}} \\
HellaSwag$^{\dagger}$ & 25.0 & 26.86 $\pm$ 0.44 & \textbf{28.00 $\pm$ 0.45} & 27.06 $\pm$ 0.44 & 27.90 $\pm$ 0.45 & \underline{27.92 $\pm$ 0.45} & 27.30 $\pm$ 0.44 \\
PIQA & 50.0 & 58.49 $\pm$ 1.15 & 59.63 $\pm$ 1.15 & \underline{60.28 $\pm$ 1.15} & \textbf{60.45 $\pm$ 1.15} & 60.07 $\pm$ 1.15 & 58.43 $\pm$ 1.16 \\
COPA & 50.0 & 54.00 $\pm$ 5.01 & 53.00 $\pm$ 5.02 & \textbf{61.00 $\pm$ 4.90} & 55.00 $\pm$ 5.00 & 55.00 $\pm$ 5.00 & \underline{56.00 $\pm$ 4.99} \\
SWAG$^{\dagger}$ & 25.0 & 36.83 $\pm$ 0.34 & 36.78 $\pm$ 0.34 & \textbf{37.26 $\pm$ 0.34} & \underline{37.09 $\pm$ 0.34} & 36.08 $\pm$ 0.34 & 34.33 $\pm$ 0.34 \\
\midrule
\multicolumn{8}{l}{\textit{WK}} \\
OpenBookQA$^{\dagger}$ & 25.0 & 28.40 $\pm$ 2.02 & 29.60 $\pm$ 2.04 & \underline{30.80 $\pm$ 2.07} & 29.60 $\pm$ 2.04 & 28.40 $\pm$ 2.02 & \textbf{31.80 $\pm$ 2.08} \\
ARC-Easy & 25.0 & 28.83 $\pm$ 0.93 & \textbf{30.22 $\pm$ 0.94} & \underline{29.42 $\pm$ 0.93} & 29.17 $\pm$ 0.94 & 28.87 $\pm$ 0.94 & 28.49 $\pm$ 0.93 \\
QA4MRE-2013$^{\dagger}$ & 20.0 & 20.77 $\pm$ 2.41 & \textbf{26.06 $\pm$ 2.61} & 21.83 $\pm$ 2.46 & 22.54 $\pm$ 2.48 & 21.83 $\pm$ 2.46 & \underline{22.89 $\pm$ 2.50} \\
\midrule
\multicolumn{8}{l}{\textit{RC}} \\
ReCoRD$^{\ddagger}$ & -- & 36.39 $\pm$ 0.48 & 37.36 $\pm$ 0.48 & 39.17 $\pm$ 0.49 & \textbf{40.93 $\pm$ 0.49} & 38.06 $\pm$ 0.48 & \underline{40.32 $\pm$ 0.49} \\
\midrule
\multicolumn{8}{l}{\textit{Gr.}} \\
BLiMP & 50.0 & \underline{74.76 $\pm$ 0.15} & 74.08 $\pm$ 0.15 & \textbf{76.37 $\pm$ 0.14} & 73.40 $\pm$ 0.15 & 72.99 $\pm$ 0.15 & 70.14 $\pm$ 0.16 \\
\midrule
\multicolumn{8}{l}{\textit{LR}} \\
LogiQA$^{\dagger}$ & 25.0 & 26.27 $\pm$ 1.73 & \textbf{28.73 $\pm$ 1.77} & \underline{28.11 $\pm$ 1.76} & 25.81 $\pm$ 1.72 & 27.50 $\pm$ 1.75 & 26.27 $\pm$ 1.73 \\
LogiQA2$^{\dagger}$ & 25.0 & 25.45 $\pm$ 1.10 & 25.06 $\pm$ 1.09 & 25.38 $\pm$ 1.10 & \textbf{26.97 $\pm$ 1.12} & \underline{25.57 $\pm$ 1.10} & 25.06 $\pm$ 1.09 \\
\bottomrule
\end{tabular}%
}
\end{table*}

\begin{table*}[h]
\centering
\small
\setlength{\tabcolsep}{4pt}
\caption{130M models final checkpoint evaluation on downstream tasks. Per-task results as presented as mean $\pm$ bootstrap standard error. Accuracy/F1 metrics are shown in \%. Section abbreviations: CS = Commonsense, WK = World Knowledge, RC = Reading Comprehension, Gr.\ = Grammar, LR = Logical Reasoning, IT = Instruction / Truthfulness. Metric symbols: $\dagger$ length-normalised accuracy, $\ddagger$ F1, $\downarrow$ lower-is-better. Per task, the best method is \textbf{bold} and the second-best is \underline{underlined}; \textit{n/a} marks tasks for which bootstrap stderr is unavailable. The Random column gives the chance-level accuracy of each task in \%; ``--'' marks tasks without a well-defined chance baseline (F1 / perplexity).}
\label{tab:eval_130m}
\resizebox{\textwidth}{!}{%
\begin{tabular}{lccccccc}
\toprule
Task & Random & \textsc{LLaMA} & \textsc{GaLore} & \textsc{Fira} & \textsc{CoLA} & \textsc{SLTrain} & \textsc{ReLoRA} \\
\midrule
\multicolumn{8}{l}{\textit{CS}} \\
HellaSwag$^{\dagger}$ & 25.0 & 28.58 $\pm$ 0.45 & \underline{29.44 $\pm$ 0.45} & \textbf{29.52 $\pm$ 0.46} & 28.78 $\pm$ 0.45 & 29.16 $\pm$ 0.45 & 26.99 $\pm$ 0.44 \\
PIQA & 50.0 & 60.28 $\pm$ 1.14 & 59.58 $\pm$ 1.14 & \underline{60.50 $\pm$ 1.14} & \textbf{60.99 $\pm$ 1.14} & 59.74 $\pm$ 1.14 & 58.81 $\pm$ 1.15 \\
COPA & 50.0 & \textbf{62.00 $\pm$ 4.88} & \underline{58.00 $\pm$ 4.96} & 58.00 $\pm$ 4.96 & 58.00 $\pm$ 4.96 & 58.00 $\pm$ 4.96 & 54.00 $\pm$ 5.01 \\
SWAG$^{\dagger}$ & 25.0 & 38.72 $\pm$ 0.34 & \textbf{39.82 $\pm$ 0.35} & \underline{39.34 $\pm$ 0.35} & 39.02 $\pm$ 0.34 & 38.90 $\pm$ 0.34 & 35.05 $\pm$ 0.34 \\
\midrule
\multicolumn{8}{l}{\textit{WK}} \\
OpenBookQA$^{\dagger}$ & 25.0 & 29.20 $\pm$ 2.04 & \textbf{32.00 $\pm$ 2.09} & 29.80 $\pm$ 2.05 & \underline{30.80 $\pm$ 2.07} & 29.40 $\pm$ 2.04 & 30.60 $\pm$ 2.06 \\
ARC-Easy & 25.0 & 30.01 $\pm$ 0.94 & \textbf{31.10 $\pm$ 0.95} & 29.50 $\pm$ 0.94 & \underline{30.68 $\pm$ 0.95} & 30.47 $\pm$ 0.94 & 28.91 $\pm$ 0.93 \\
QA4MRE-2013$^{\dagger}$ & 20.0 & 23.24 $\pm$ 2.51 & 21.48 $\pm$ 2.44 & 24.30 $\pm$ 2.55 & 19.37 $\pm$ 2.35 & \underline{25.00 $\pm$ 2.57} & \textbf{25.35 $\pm$ 2.59} \\
\midrule
\multicolumn{8}{l}{\textit{RC}} \\
ReCoRD$^{\ddagger}$ & -- & 43.98 $\pm$ 0.49 & 39.00 $\pm$ 0.49 & \underline{45.86 $\pm$ 0.50} & \textbf{49.60 $\pm$ 0.50} & 42.48 $\pm$ 0.49 & 39.02 $\pm$ 0.49 \\
\midrule
\multicolumn{8}{l}{\textit{Gr.}} \\
BLiMP & 50.0 & 75.96 $\pm$ 0.14 & 76.00 $\pm$ 0.14 & \underline{77.39 $\pm$ 0.14} & \textbf{78.19 $\pm$ 0.14} & 76.71 $\pm$ 0.14 & 74.62 $\pm$ 0.15 \\
\midrule
\multicolumn{8}{l}{\textit{LR}} \\
LogiQA$^{\dagger}$ & 25.0 & \underline{28.26 $\pm$ 1.77} & 25.65 $\pm$ 1.71 & 27.19 $\pm$ 1.75 & \textbf{28.88 $\pm$ 1.78} & 26.11 $\pm$ 1.72 & 26.73 $\pm$ 1.74 \\
LogiQA2$^{\dagger}$ & 25.0 & \textbf{27.61 $\pm$ 1.13} & 27.61 $\pm$ 1.13 & 26.78 $\pm$ 1.12 & 26.65 $\pm$ 1.12 & 25.64 $\pm$ 1.10 & 26.08 $\pm$ 1.11 \\
\bottomrule
\end{tabular}%
}
\end{table*}

\begin{table*}[h]
\centering
\small
\setlength{\tabcolsep}{4pt}
\caption{350M models final checkpoint evaluation on downstream tasks. Per-task results as presented as mean $\pm$ bootstrap standard error. Accuracy/F1 metrics are shown in \%. Section abbreviations: CS = Commonsense, WK = World Knowledge, RC = Reading Comprehension, Gr.\ = Grammar, LR = Logical Reasoning, IT = Instruction / Truthfulness. Metric symbols: $\dagger$ length-normalised accuracy, $\ddagger$ F1, $\downarrow$ lower-is-better. Per task, the best method is \textbf{bold} and the second-best is \underline{underlined}; \textit{n/a} marks tasks for which bootstrap stderr is unavailable. The Random column gives the chance-level accuracy of each task in \%; ``--'' marks tasks without a well-defined chance baseline (F1 / perplexity).}
\label{tab:eval_350m}
\resizebox{\textwidth}{!}{%
\begin{tabular}{lccccccc}
\toprule
Task & Random & \textsc{LLaMA} & \textsc{GaLore} & \textsc{Fira} & \textsc{CoLA} & \textsc{SLTrain} & \textsc{ReLoRA} \\
\midrule
\multicolumn{8}{l}{\textit{CS}} \\
HellaSwag$^{\dagger}$ & 25.0 & 31.75 $\pm$ 0.46 & 31.95 $\pm$ 0.47 & \textbf{34.44 $\pm$ 0.47} & \underline{32.41 $\pm$ 0.47} & 31.23 $\pm$ 0.46 & 27.59 $\pm$ 0.45 \\
PIQA & 50.0 & 61.81 $\pm$ 1.13 & \underline{63.17 $\pm$ 1.13} & \textbf{63.82 $\pm$ 1.12} & 61.75 $\pm$ 1.13 & 62.19 $\pm$ 1.13 & 59.41 $\pm$ 1.15 \\
COPA & 50.0 & \textbf{66.00 $\pm$ 4.76} & 60.00 $\pm$ 4.92 & 63.00 $\pm$ 4.85 & \underline{65.00 $\pm$ 4.79} & 61.00 $\pm$ 4.90 & 60.00 $\pm$ 4.92 \\
SWAG$^{\dagger}$ & 25.0 & 41.19 $\pm$ 0.35 & 42.37 $\pm$ 0.35 & \textbf{42.67 $\pm$ 0.35} & \underline{42.65 $\pm$ 0.35} & 41.79 $\pm$ 0.35 & 35.79 $\pm$ 0.34 \\
\midrule
\multicolumn{8}{l}{\textit{WK}} \\
OpenBookQA$^{\dagger}$ & 25.0 & 29.20 $\pm$ 2.04 & 34.00 $\pm$ 2.12 & \textbf{36.40 $\pm$ 2.15} & \underline{34.40 $\pm$ 2.13} & 32.20 $\pm$ 2.09 & 29.80 $\pm$ 2.05 \\
ARC-Easy & 25.0 & \underline{31.06 $\pm$ 0.95} & \textbf{31.48 $\pm$ 0.95} & 30.85 $\pm$ 0.95 & 29.80 $\pm$ 0.94 & 30.51 $\pm$ 0.94 & 29.00 $\pm$ 0.93 \\
QA4MRE-2013$^{\dagger}$ & 20.0 & 23.59 $\pm$ 2.52 & 23.24 $\pm$ 2.51 & \underline{23.94 $\pm$ 2.54} & 22.54 $\pm$ 2.48 & \textbf{25.35 $\pm$ 2.59} & 17.25 $\pm$ 2.25 \\
\midrule
\multicolumn{8}{l}{\textit{RC}} \\
ReCoRD$^{\ddagger}$ & -- & 45.41 $\pm$ 0.50 & 41.16 $\pm$ 0.49 & 47.00 $\pm$ 0.50 & \textbf{55.54 $\pm$ 0.49} & \underline{52.16 $\pm$ 0.50} & 42.37 $\pm$ 0.49 \\
\midrule
\multicolumn{8}{l}{\textit{Gr.}} \\
BLiMP & 50.0 & \textbf{79.46 $\pm$ 0.14} & 77.06 $\pm$ 0.14 & 76.20 $\pm$ 0.15 & 78.16 $\pm$ 0.14 & \underline{78.70 $\pm$ 0.14} & 76.55 $\pm$ 0.15 \\
\midrule
\multicolumn{8}{l}{\textit{LR}} \\
LogiQA$^{\dagger}$ & 25.0 & 26.11 $\pm$ 1.72 & 27.65 $\pm$ 1.75 & 25.50 $\pm$ 1.71 & \textbf{29.03 $\pm$ 1.78} & \underline{27.80 $\pm$ 1.76} & 27.34 $\pm$ 1.75 \\
LogiQA2$^{\dagger}$ & 25.0 & 25.89 $\pm$ 1.11 & 26.27 $\pm$ 1.11 & \underline{27.04 $\pm$ 1.12} & 26.65 $\pm$ 1.12 & \textbf{27.93 $\pm$ 1.13} & 25.64 $\pm$ 1.10 \\
\bottomrule
\end{tabular}%
}
\end{table*}

\section{More results}
\label{appendix:more_results}
We report additional plots and results for all the metric families in this section.

\subsection{1-D loss landscape}
\label{appendix:more_results_1D_loss_landscape}
This section reports the full set of 1-D loss landscape plots complementing the 350M random-direction results and the top-1 PCA results in Figure ~\ref{fig:interp_combined} of the main paper. Figures~\ref{fig:landscape-60m-all-wo-relora}--\ref{fig:landscape-350m-all} show random-direction landscapes at all three scales (60M, 130M, 350M), each presented twice --- once excluding ReLoRA and once including it. Because ReLoRA's sharpness is one to two orders of magnitude larger than the other methods at every scale, including it compresses the $y$-axis and obscures the relative ordering among the remaining methods; we therefore provide both views. Figures~\ref{fig:pca-k20-60m}--\ref{fig:pca-k20-350m} extend the top-$k$ PCA-direction analysis to $k \in \{1, 5, 10, 20\}$ at each scale.

\begin{figure}[t]
    \centering
    \includegraphics[width=1\linewidth]{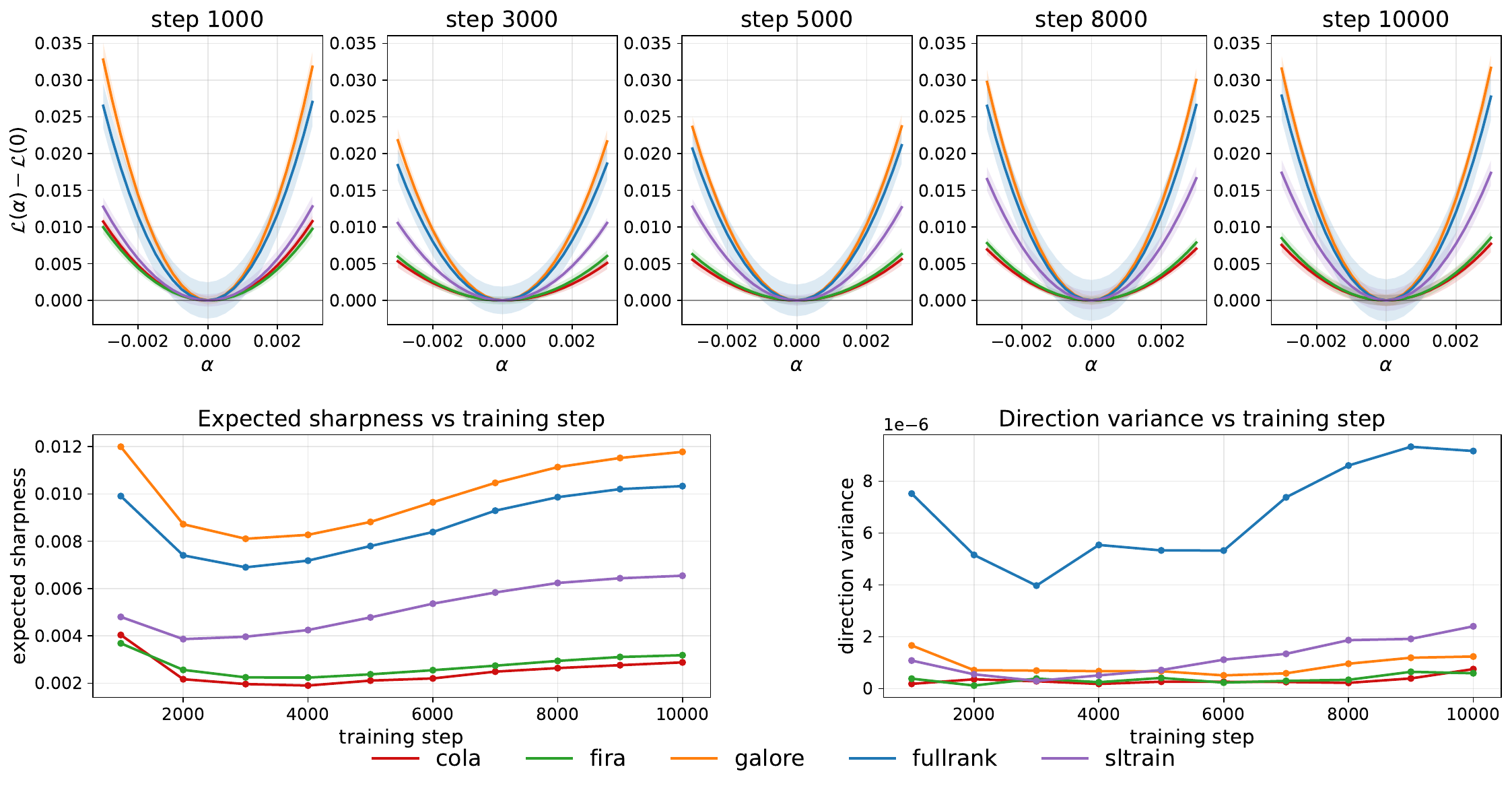}
    \caption{\textbf{1-D loss landscape at 60M parameters.}
    \emph{Top:} centered loss profile $\mathcal{L}(\alpha)-\mathcal{L}(0)$
    averaged over 100 random directions, at five training checkpoints
    (1k, 3k, 5k, 8k, 10k).
    \emph{Bottom-left:} average sharpness with respect to training step.
    \emph{Bottom-right:} average direction variance.
    \textbf{ReLoRA is omitted} from the plot as it makes it harder to view other methods. The plot including ReLoRA is given in ~\ref{fig:landscape-60m-all}}.
    \label{fig:landscape-60m-all-wo-relora}
\end{figure}

\begin{figure}[t]
    \centering
    \includegraphics[width=1\linewidth]{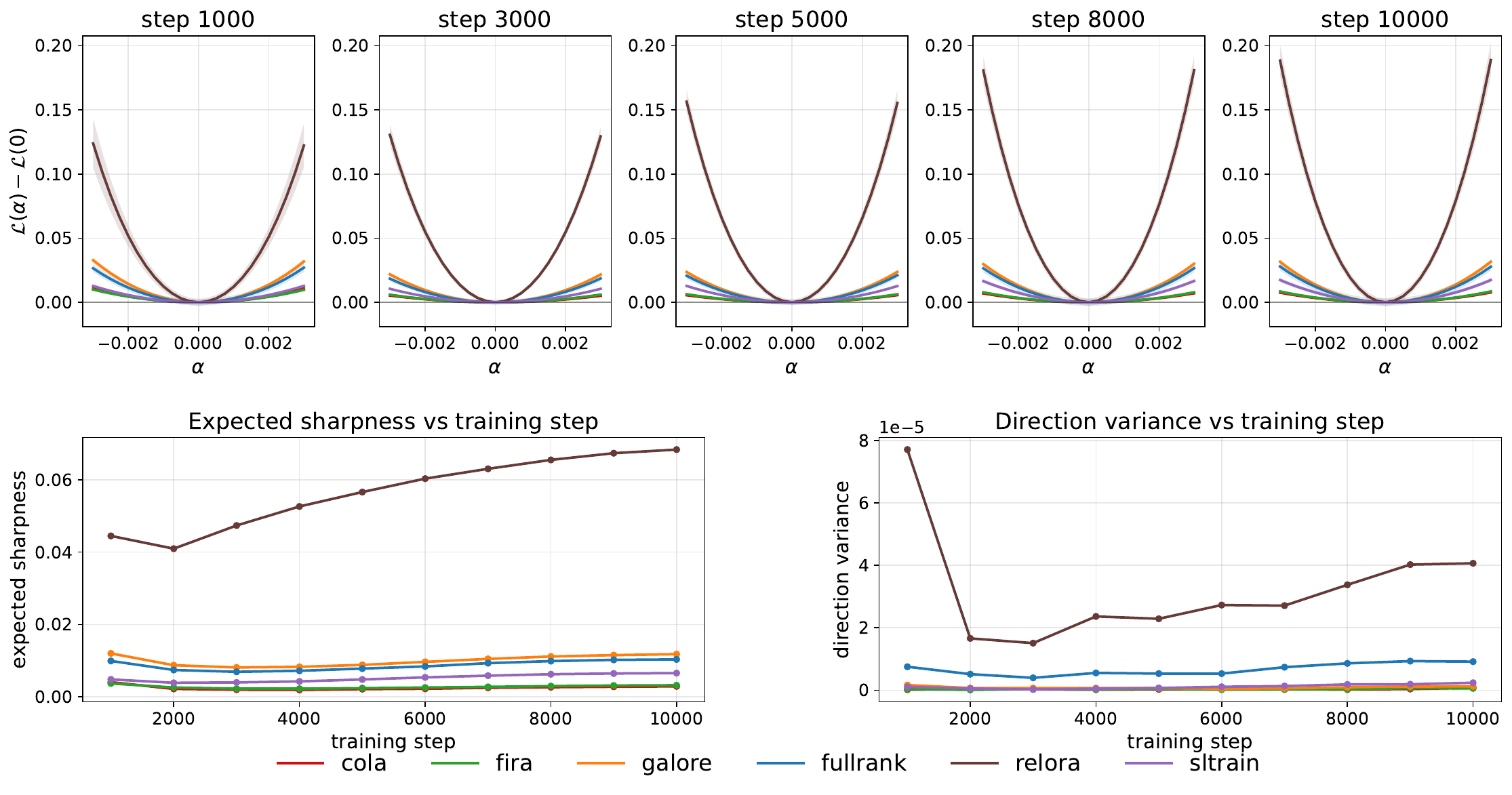}
    \caption{\textbf{1-D loss landscape at 60M parameters for \textbf{All methods}.}
    \emph{Top:} centered loss profile $\mathcal{L}(\alpha)-\mathcal{L}(0)$
    averaged over 100 random directions, at five training checkpoints
    (1k, 3k, 5k, 8k, 10k).
    \emph{Bottom-left:} average expected sharpness with respect to training step.
    \emph{Bottom-right:} average direction variance.}
    \label{fig:landscape-60m-all}
\end{figure}

\begin{figure}[t]
    \centering
    \includegraphics[width=1\linewidth]{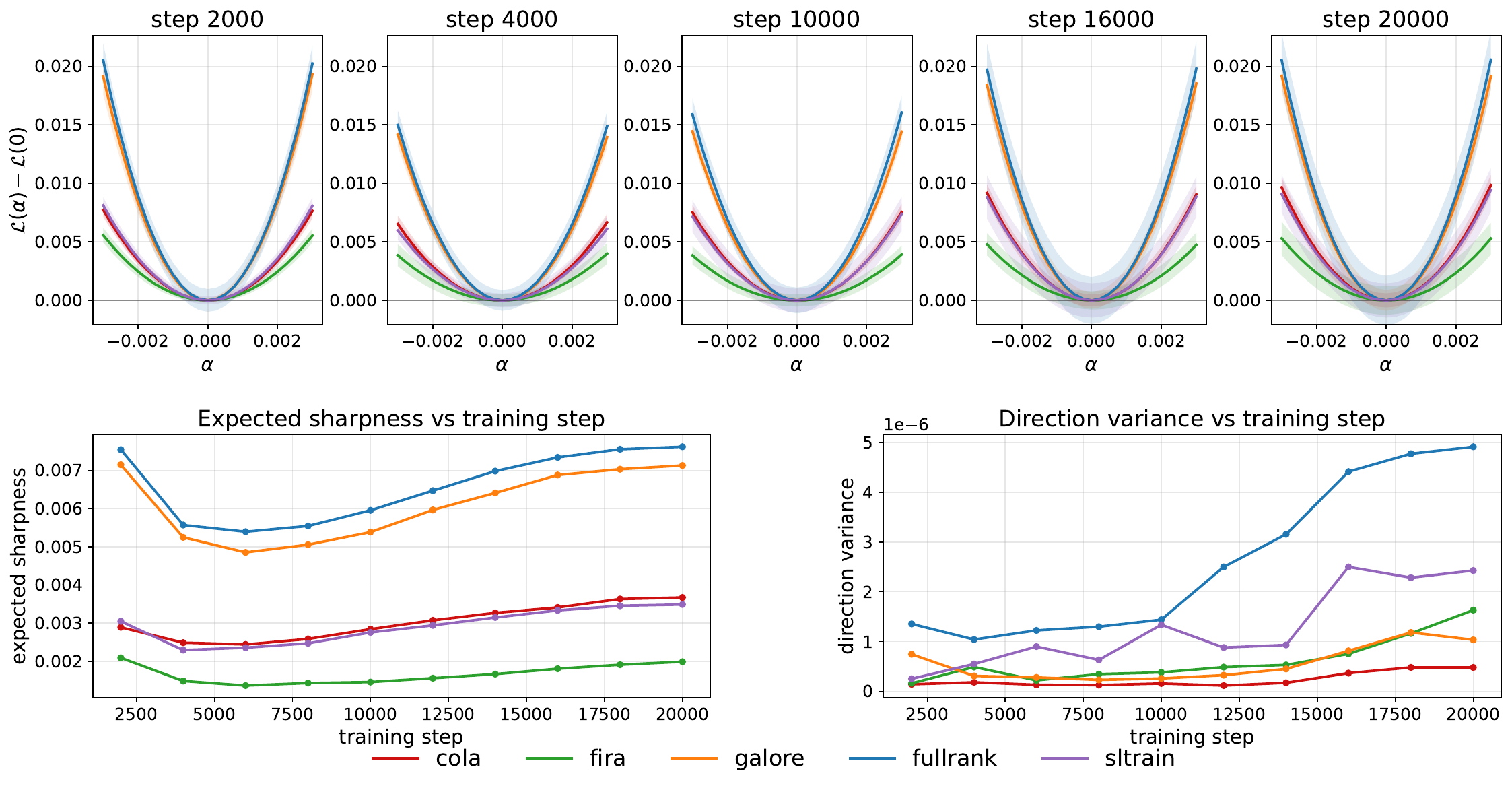}
    \caption{\textbf{1-D loss landscape at 130M parameters.}
    Same layout as Figure \ref{fig:landscape-60m-all-wo-relora}, with top-row checkpoints at
    $\{2\text{k}, 6\text{k}, 10\text{k}, 14\text{k}, 20\text{k}\}$.}
    \label{fig:landscape-130m-all-wo-relora}
\end{figure}

\begin{figure}[t]
    \centering
    \includegraphics[width=1\linewidth]{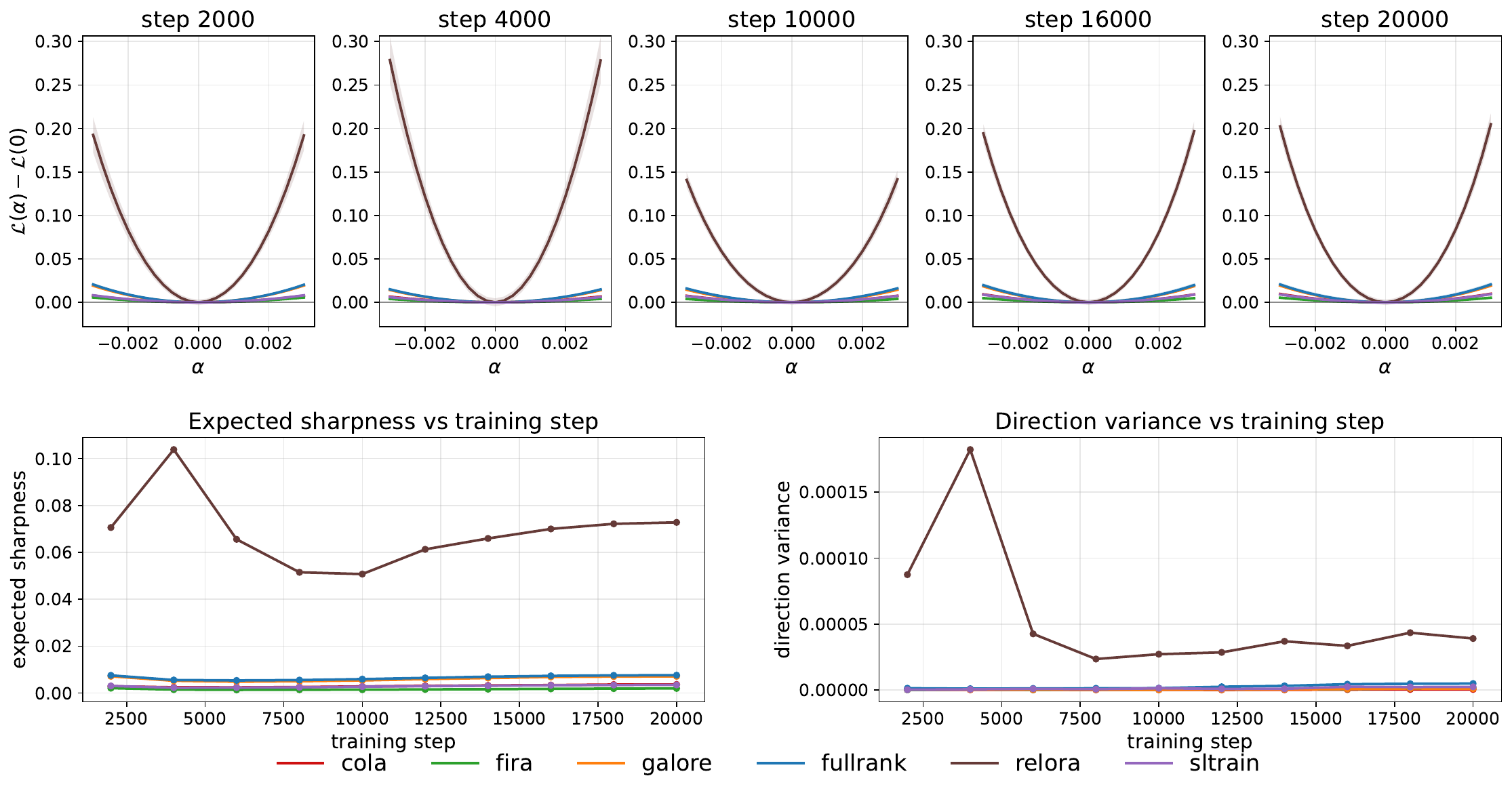}
    \caption{1-D loss landscape at 130M parameters for \textbf{All methods}.}
    \label{fig:landscape-130m-all}
\end{figure}

\begin{figure}[t]
    \centering
    \includegraphics[width=1\linewidth]{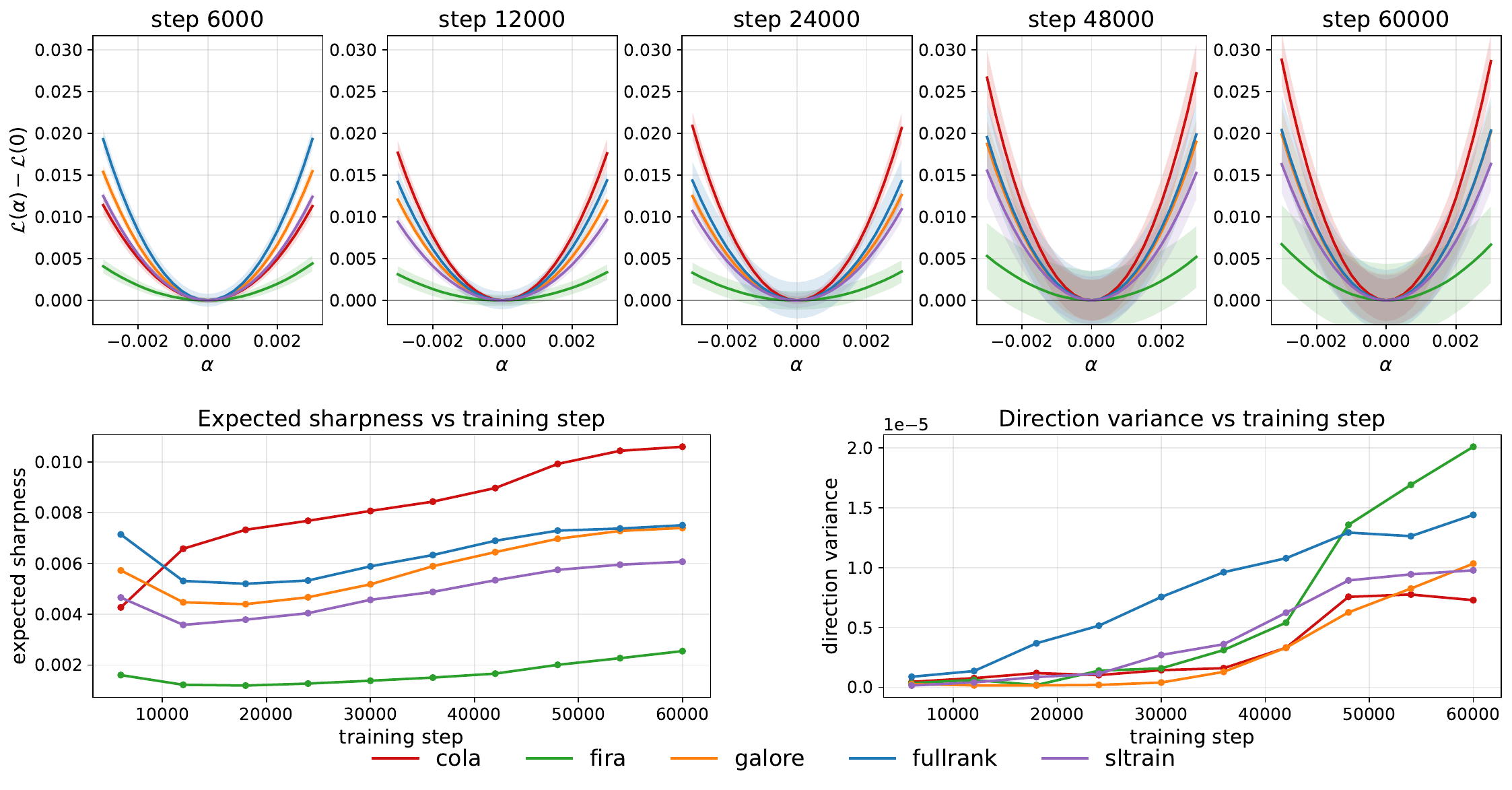}
    \caption{\textbf{1-D loss landscape at 350M parameters.}
    Same layout as \ref{fig:landscape-60m-all-wo-relora}.}
    \label{fig:landscape-350m-all-wo-relora}
\end{figure}

\begin{figure}[t]
    \centering
    \includegraphics[width=1\linewidth]{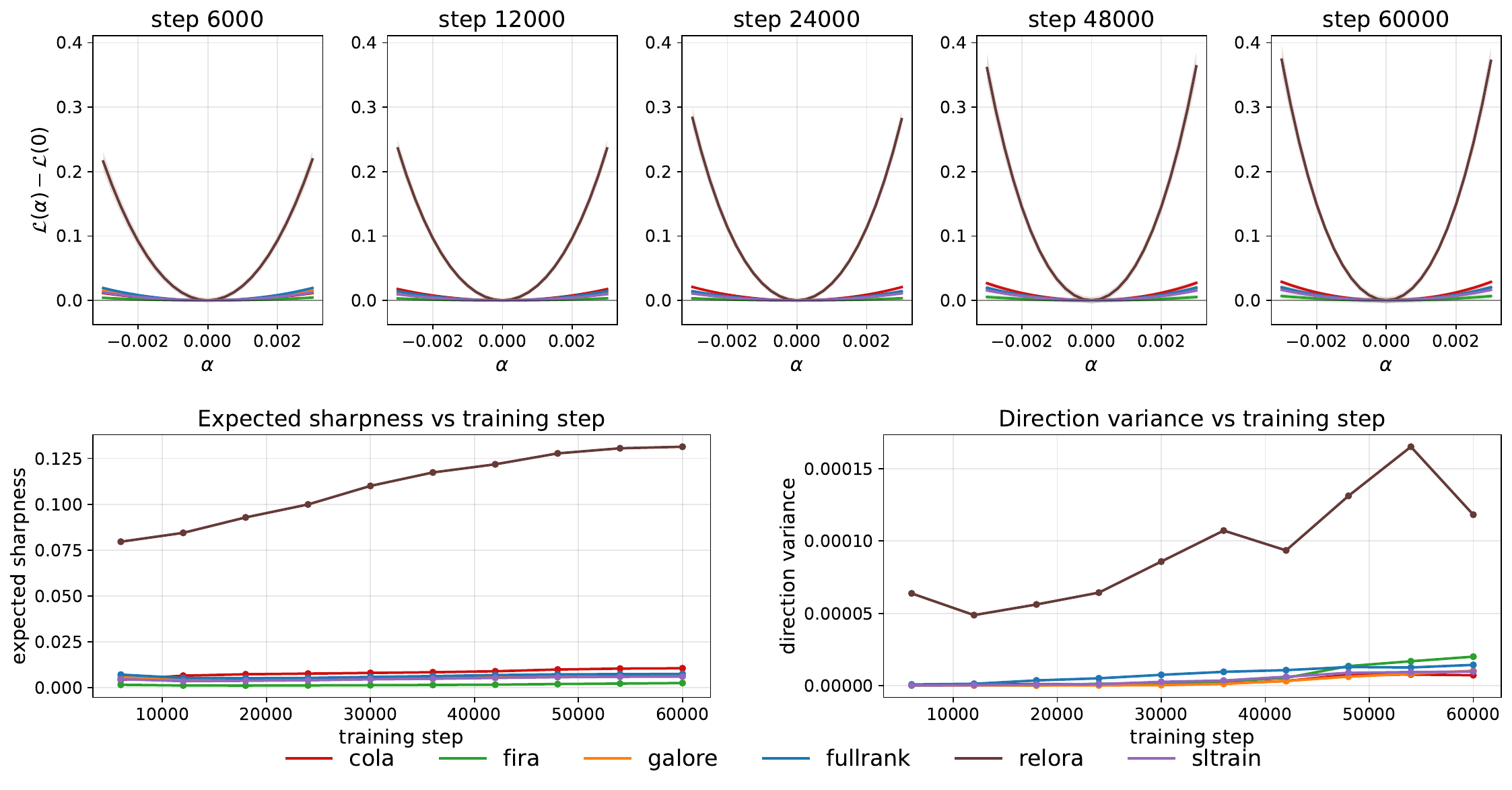}
    \caption{1-D loss landscape at 350M parameters for \textbf{All methods}.}
    \label{fig:landscape-350m-all}
\end{figure}


\begin{figure}[h]
    \centering
    \includegraphics[width=\linewidth]{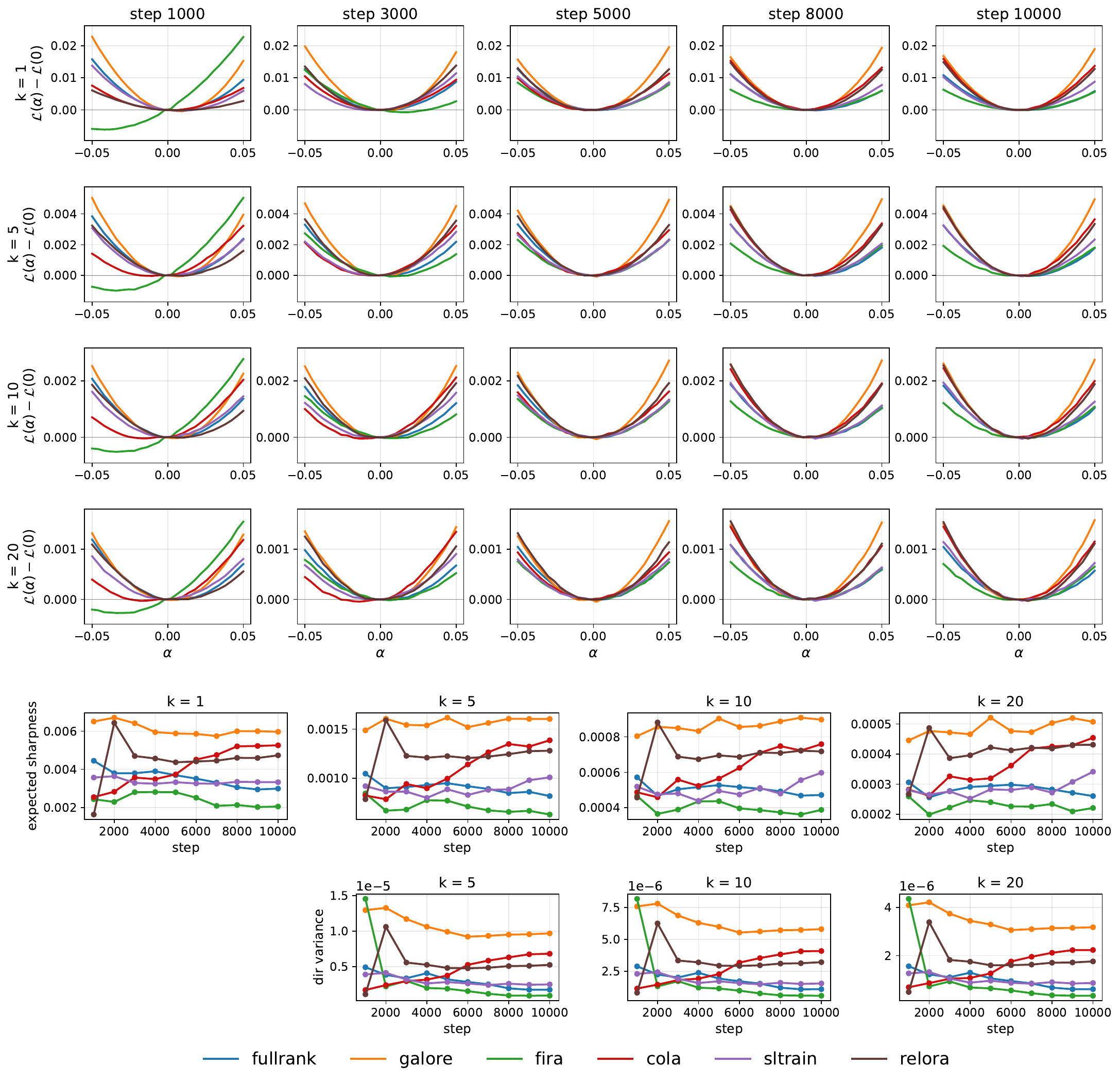}
    \caption{\textbf{1-D loss landscape along top-$k$ PCA directions at 60M parameters} for $k \in \{1, 5, 10, 20\}$. \emph{Top four rows:} centered loss profile at five training checkpoints. \emph{Bottom:} expected sharpness (left column of summary panels) and across-component direction variance (right column) as a function of training step, per $k$.}
    \label{fig:pca-k20-60m}
\end{figure}

\begin{figure}[h]
    \centering
    \includegraphics[width=\linewidth]{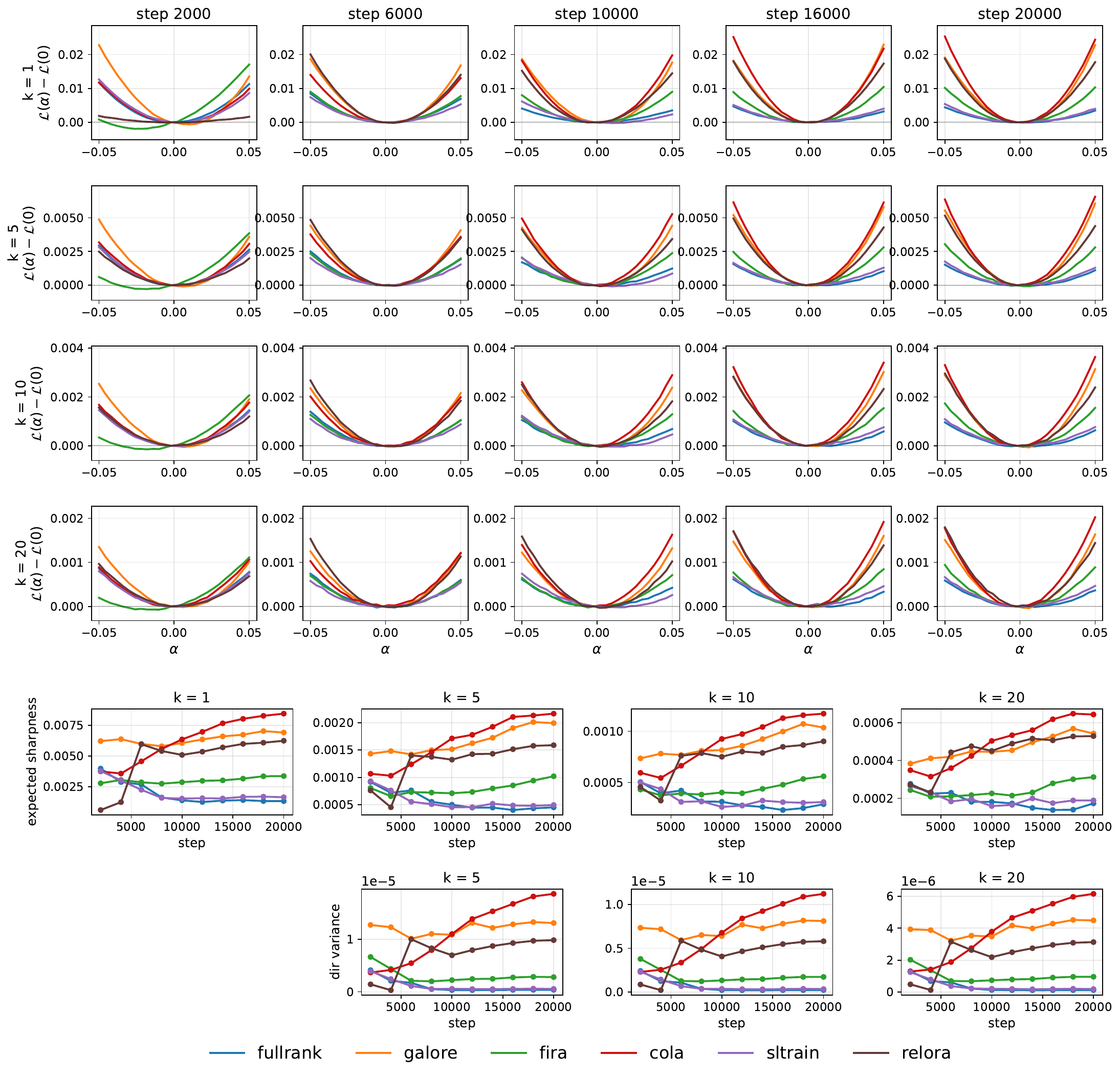}
    \caption{\textbf{1-D loss landscape along top-$k$ PCA directions at 130M parameters} for $k \in \{1, 5, 10, 20\}$. Layout identical to Figure~\ref{fig:pca-k20-60m}.}
    \label{fig:pca-k20-130m}
\end{figure}

\begin{figure}[h]
    \centering
    \includegraphics[width=\linewidth]{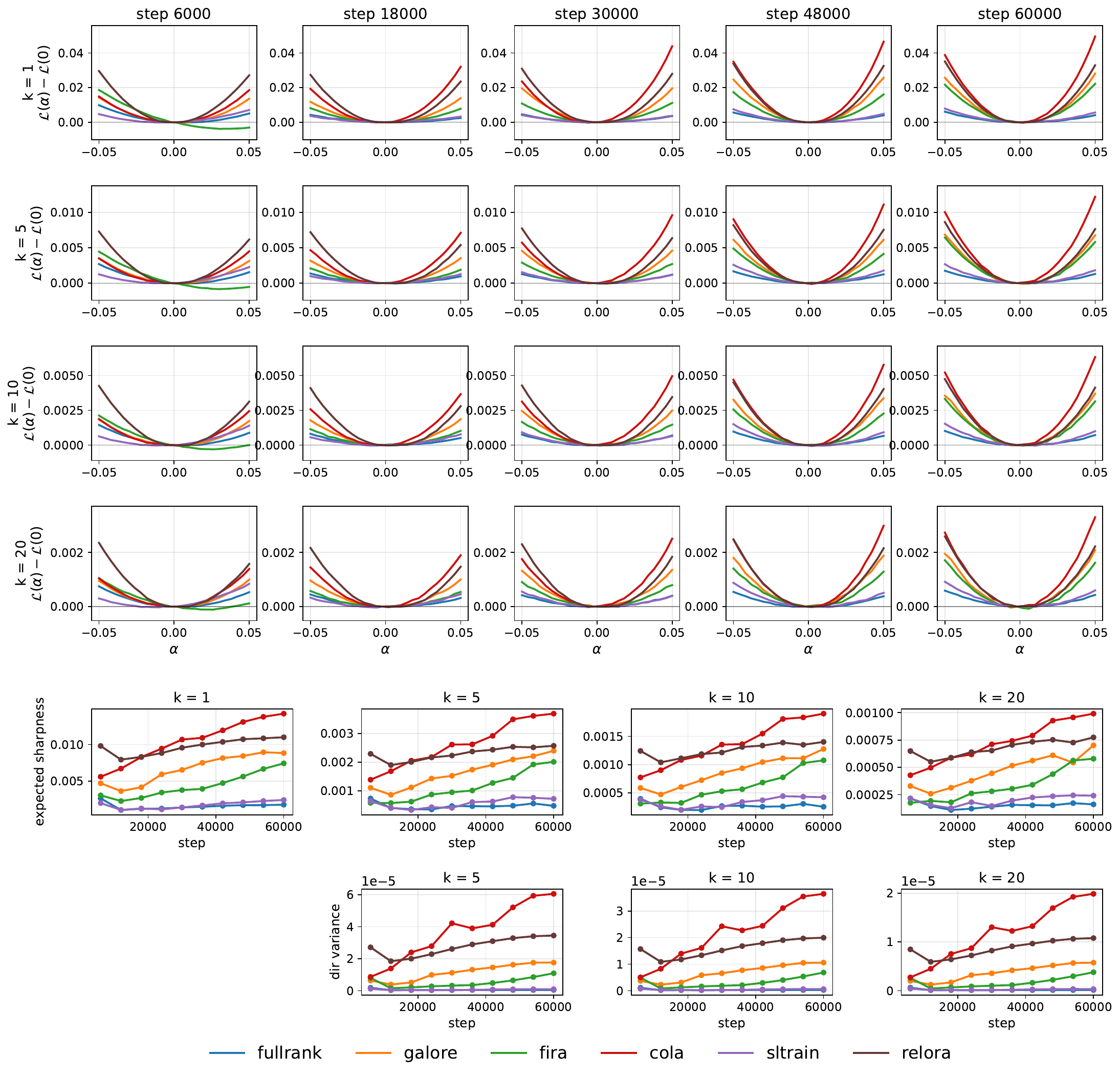}
    \caption{\textbf{1-D loss landscape along top-$k$ PCA directions at 350M parameters} for $k \in \{1, 5, 10, 20\}$. Layout identical to Figure~\ref{fig:pca-k20-60m}. CoLA's direction variance grows monotonically with training across all $k$, while other methods remain near zero, indicating that CoLA's sharpness is distributed anisotropically across its top-$k$ subspace rather than concentrated in any single direction.}
    \label{fig:pca-k20-350m}
\end{figure}

\section{Rank related metrics}
\label{appendix:results_rank}
This section reports the rank and spectral metric trajectories at the 60M and 130M scales, complementing the 350M results shown in Figure ~\ref{fig:composite_rank} of the main paper. Both figures follow the same layout as Figure~\ref{fig:composite_rank} \emph{Row 1} tracks the four spectral metrics --- effective rank, stable rank, spectral gap, and the count of singular values above $0.1$ --- on the trained weight $W$ across training; \emph{Row 2} tracks the same metrics on the consecutive-checkpoint update $\Delta W = W_{t+1} - W_{t}$; and \emph{Row 3} shows the singular value distributions of $\Delta W$ per projection type ($W_Q$, $W_V$, $W_{\text{up}}$), with the rightmost panel reporting the per-projection count of singular values above $0.1$. The same method ordering observed at 350M --- GaLore $\approx$ Fira $>$ SLTrain $>$ CoLA $>$ ReLoRA --- holds at both smaller scales: optimizer-based methods (GaLore, Fira) most faithfully reproduce full-rank's update geometry, while parameterization-based methods (CoLA, SLTrain) and ReLoRA systematically underpopulate the small-singular-value tail in the attention $W_V$/$W_O$ blocks.

\begin{figure}
    \centering
    \includegraphics[width=\linewidth]{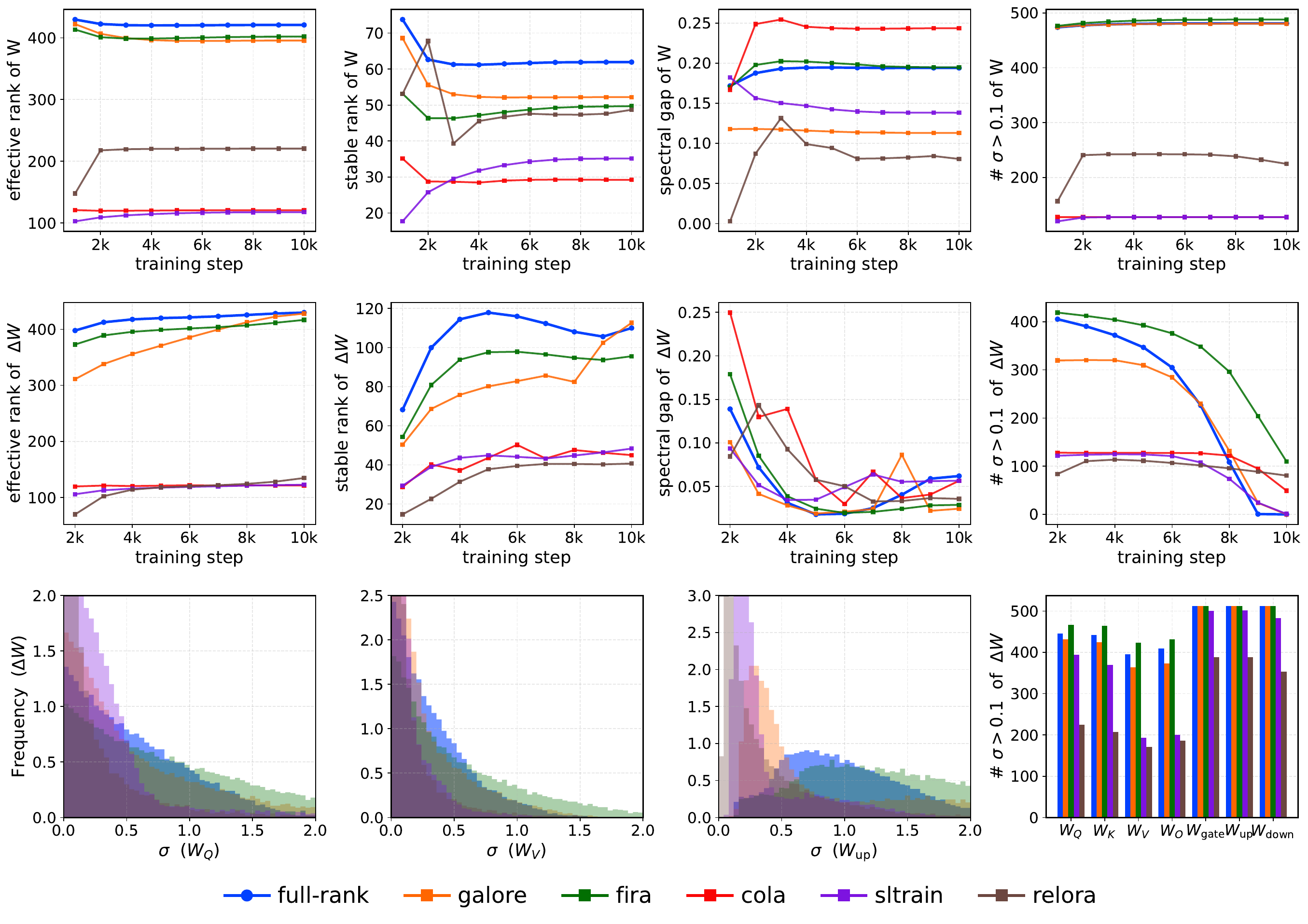}
    \caption{Rank and Spectral metrics for 60M.}
    \label{fig:appendix_rank_60m}
\end{figure}

\begin{figure}
    \centering
    \includegraphics[width=\linewidth]{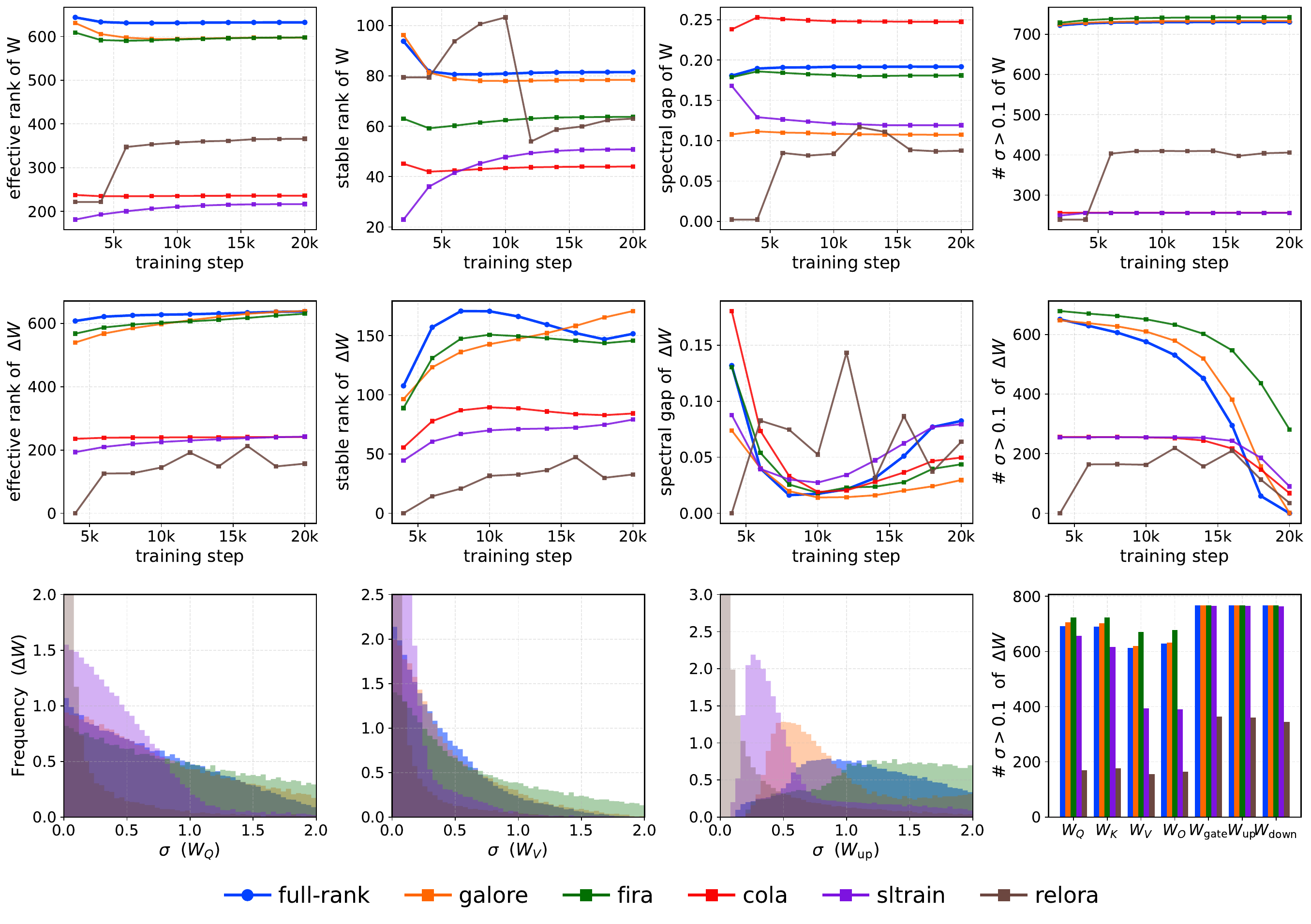}
    \caption{Rank and Spectral metrics for 130M.}
    \label{fig:appendix_rank_130m}
\end{figure}

\section{Activations}
\label{appendix:results_activation_full}
This section reports the full per-layer $\times$ per-training-step activation heatmaps for every method at every model scale, complementing the bar summaries and the Fira/CoLA-only heatmaps shown in Figure ~\ref{fig:composite_activation} of the main paper. All three figures share the same grid layout: rows index the five low-rank methods (GaLore, Fira, CoLA, SLTrain, ReLoRA) and columns index the three model scales (60M, 130M, 350M); within each panel, the $x$-axis is the training step and the $y$-axis is the decoder layer index, where layer 0 corresponds to the embedding output. Figure~\ref{fig:act-last-l2-layers} reports the per-layer Activation L2 distance to the full-rank baseline (Eq.~\ref{eq:act-l2}), Figure~\ref{fig:act-cka-layers} reports the per-layer Linear CKA similarity (Eq.~\ref{eq:act-cka}), and Figure~\ref{fig:act-cosine-layers} reports the per-layer Activation cosine similarity (Eq.~\ref{eq:act-cos}). Across all three metrics, divergence from full-rank concentrates in the later decoder layers and grows with training, with CoLA showing the most severe directional mismatch --- cosine similarity remains near zero across all layers and steps at 130M and 350M.

\begin{figure}
    \centering
    \includegraphics[width=1.1\linewidth]{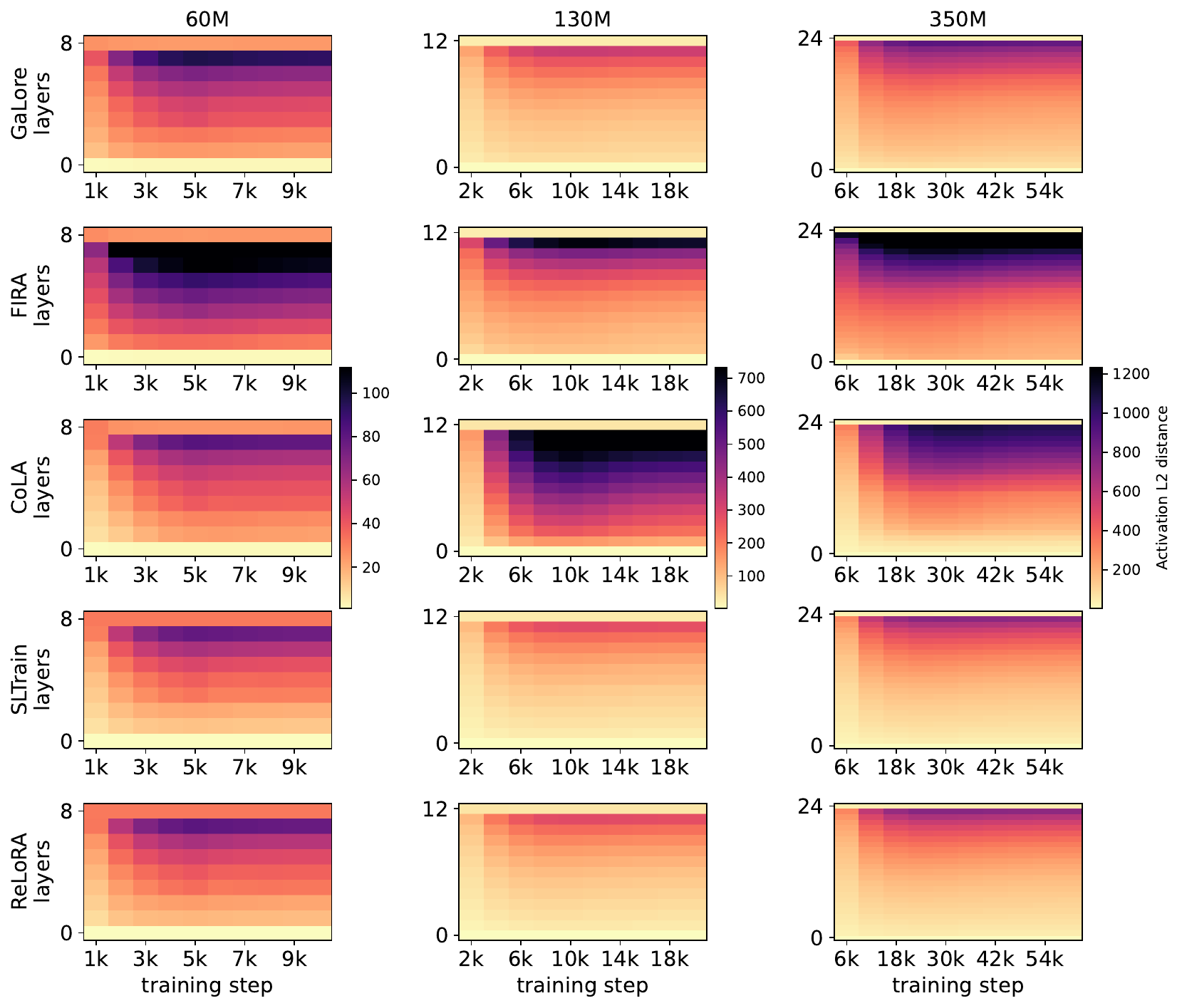}
    \caption{Activation L2 distance layer-wise}
    \label{fig:act-last-l2-layers}
\end{figure}

\begin{figure}
    \centering
    \includegraphics[width=1.1\linewidth]{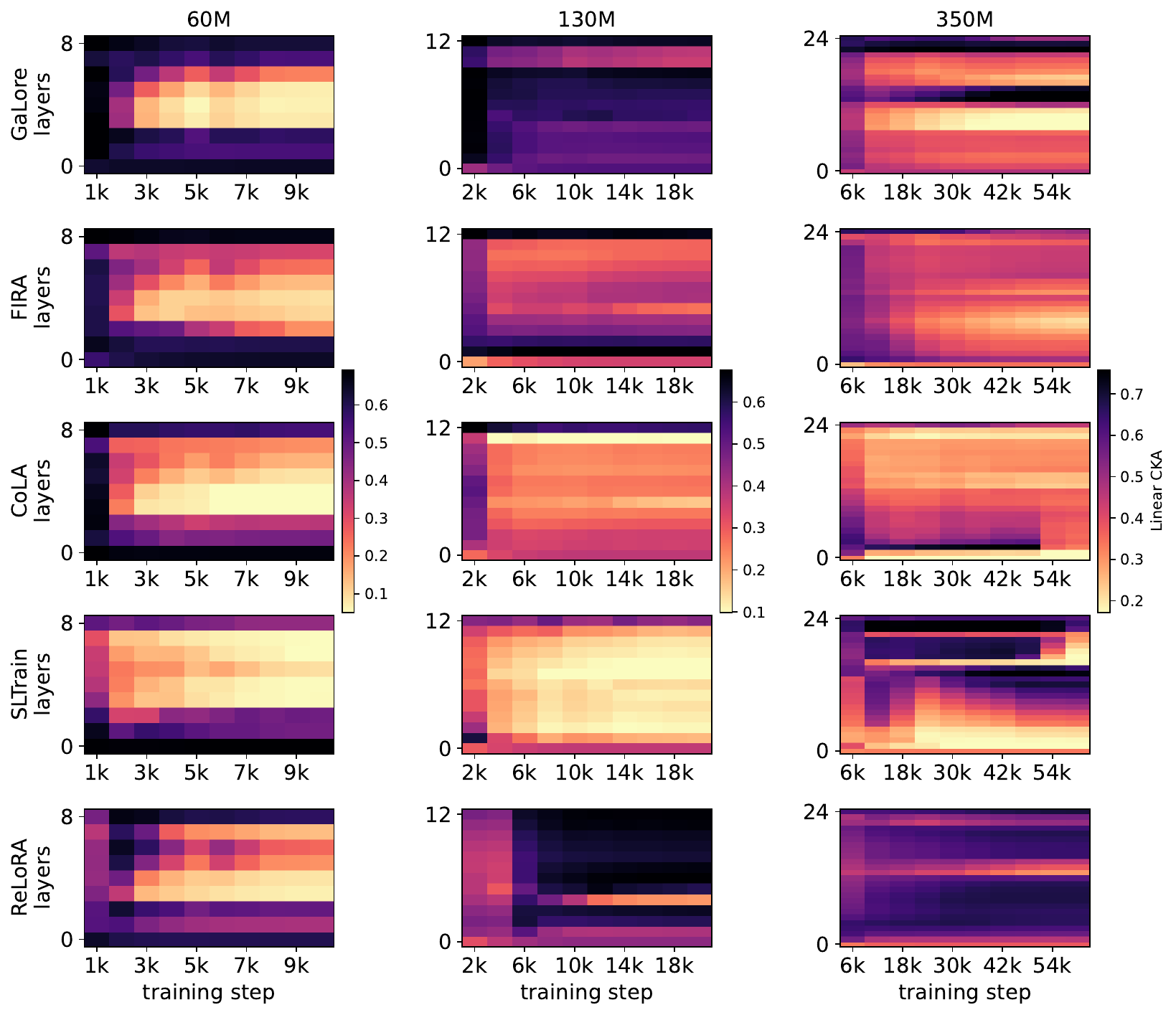}
    \caption{Activation linear CKA similarity}
    \label{fig:act-cka-layers}
\end{figure}

\begin{figure}
    \centering
    \includegraphics[width=1.1\linewidth]{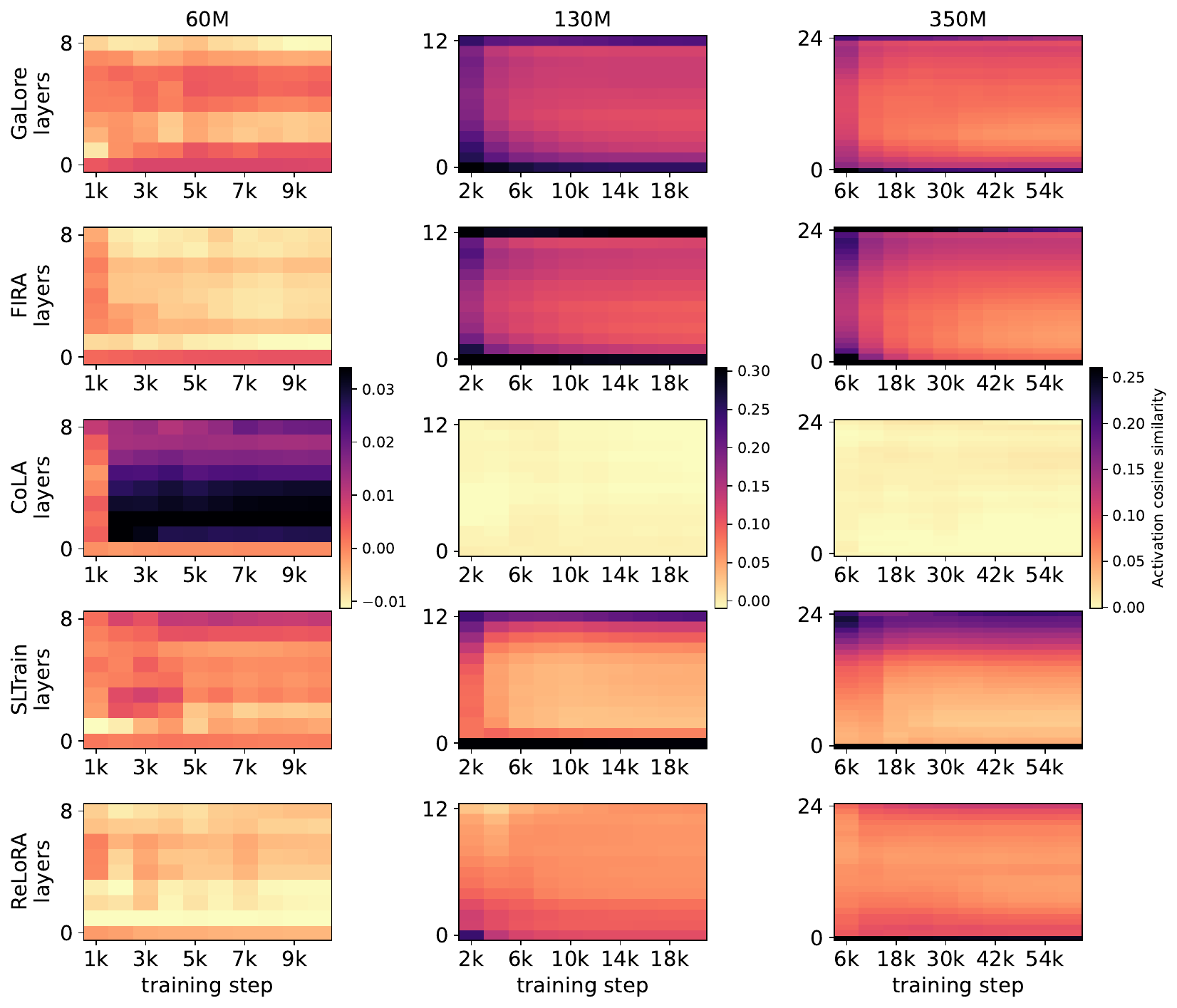}
    \caption{Activation cosine similarity layer-wise}
    \label{fig:act-cosine-layers}
\end{figure}

\newpage
\section{Correlation results}


\begin{table}[t]
\centering
\caption{Top-8 sign-consistent geometry features used in the combined predictor. Per-size Spearman correlation between each feature and the 11-task downstream mean. All three sizes share the same sign; features are ranked by $|\text{median}\,\rho|$.}
\label{tab:top8_features}
\small
\begin{tabular}{llccccc}
\toprule
\# & Feature & Lens & 60M & 130M & 350M & $|\widetilde{\rho}|$ \\
\midrule
1 & \texttt{barrier\_consec} (CCBH) & interpolation       & $-0.69$ & $-0.70$ & $-0.55$ & 0.69 \\
2 & \texttt{act\_cka\_mean}         & activation          & $-0.62$ & $-0.42$ & $-0.78$ & 0.62 \\
3 & \texttt{act\_l2\_mean}          & activation          & $+0.66$ & $+0.62$ & $+0.55$ & 0.62 \\
4 & \texttt{stable\_rank\_dW}       & rank $\Delta W$     & $+0.28$ & $+0.62$ & $+0.35$ & 0.35 \\
5 & \texttt{eff\_rank\_dW}          & rank $\Delta W$     & $+0.30$ & $+0.70$ & $+0.18$ & 0.30 \\
6 & \texttt{threshold\_rank\_W}     & rank $W$            & $+0.21$ & $+0.37$ & $+0.05$ & 0.21 \\
7 & \texttt{spectral\_gap\_dW}      & rank $\Delta W$     & $-0.29$ & $-0.20$ & $-0.18$ & 0.20 \\
8 & \texttt{stable\_rank\_W}        & rank $W$            & $-0.20$ & $-0.11$ & $-0.17$ & 0.17 \\
\bottomrule
\end{tabular}
\end{table}

\end{document}